\documentclass{article}

\usepackage{microtype}
\usepackage{graphicx}
\usepackage{booktabs}
\usepackage{graphicx}
\usepackage{multicol}
\usepackage{amssymb}
\usepackage{amsmath}
\usepackage{amsthm}
\usepackage{xcolor}
\usepackage{subcaption}
\usepackage{wrapfig}
\usepackage{xspace} 
\usepackage{enumitem}
\usepackage{hyperref}
\usepackage{algorithm}
\usepackage{algorithmic}

\newcommand{\fedavg}{\texttt{FedAvg}\xspace}
\newcommand{\fedprox}{\texttt{FedProx}\xspace}
\newcommand{\feddane}{\texttt{FedDane}\xspace}

\newcommand{\bE}{{\mathbb{E}}}
\newcommand{\E}[2]{\mathbb{E}_{#1}{\left[{#2}\right]}}

\newcommand{\eqdef}{\mathrel{\mathop:}=}

\theoremstyle{plain}
\newtheorem{theorem}{Theorem}

\newtheorem{assumption}{Assumption}
\newtheorem{remark}[theorem]{Remark}
\newtheorem{corollary}[theorem]{Corollary}
\theoremstyle{definition}
\newtheorem{definition}[theorem]{Definition}

\usepackage[accepted]{mlsys2020}

\mlsystitlerunning{Federated Optimization in Heterogeneous Networks}

\begin{document}

\twocolumn[
\mlsystitle{Federated Optimization in Heterogeneous Networks}



\mlsyssetsymbol{equal}{*}

\begin{mlsysauthorlist}
\mlsysauthor{Tian Li}{cmu}
\mlsysauthor{Anit Kumar Sahu}{bcai}
\mlsysauthor{Manzil Zaheer}{google}
\mlsysauthor{Maziar Sanjabi}{fb}
\mlsysauthor{Ameet Talwalkar}{cmu,determined}
\mlsysauthor{Virginia Smith}{cmu}
\end{mlsysauthorlist}

\mlsysaffiliation{cmu}{Carnegie Mellon University}
\mlsysaffiliation{bcai}{Bosch Center for Artificial Intelligence}
\mlsysaffiliation{google}{Goolge Research}
\mlsysaffiliation{fb}{Facebook AI}
\mlsysaffiliation{determined}{Determined AI}

\mlsyscorrespondingauthor{Tian Li}{tianli@cmu.edu}

\mlsyskeywords{}

\vskip 0.3in

\begin{abstract}
Federated Learning is a distributed learning paradigm with two key challenges that differentiate it from traditional distributed optimization: (1) significant variability in terms of the systems characteristics on each device in the network (systems heterogeneity), and (2) non-identically distributed data across the network (statistical heterogeneity). 
In this work, we introduce a framework, \fedprox, to tackle heterogeneity in federated networks. \fedprox can be viewed as a generalization and re-parametrization of \fedavg, the current state-of-the-art method for federated learning. While this re-parameterization makes only minor modifications to the method itself, these modifications have important ramifications both in theory and in practice. 
Theoretically, we provide convergence guarantees for our framework when learning over data from non-identical distributions (statistical heterogeneity), and while adhering to  device-level systems constraints by allowing each participating device to perform a variable amount of work (systems heterogeneity). 
Practically, we demonstrate that \fedprox allows for more robust convergence than \fedavg across a suite of realistic federated datasets. In particular, in highly heterogeneous settings, \fedprox demonstrates significantly more stable and accurate convergence behavior relative to \fedavg---improving absolute test accuracy by 22\% on average. 
\end{abstract}
]

\printAffiliationsAndNotice{}

\section{Introduction}
\label{sec:intro}

Federated learning has emerged as an attractive paradigm for distributing training of machine learning models in networks of remote devices. While there is a wealth of work on distributed optimization in the context of machine learning, two key challenges distinguish federated learning from traditional distributed optimization: high degrees of \emph{systems and statistical heterogeneity}\footnote{Privacy is a third key challenge in the federated setting.  While not the focus of this work, standard privacy-preserving approaches such as differential privacy and secure multiparty communication can naturally be combined with the methods proposed herein---particularly since our framework proposes only lightweight algorithmic modifications to prior work.}~\citep{mcmahan2016FedAvg,lireviewpaper}.

In an attempt to handle heterogeneity and tackle high communication costs, optimization methods that allow for local updating and low participation are a popular approach for federated learning~\citep{mcmahan2016FedAvg,fed_multitask_smith_2017}. 
In particular, \fedavg~\citep{mcmahan2016FedAvg} is an iterative method that has emerged as the de facto optimization method in the federated setting. At each iteration, \fedavg first locally performs $E$ epochs of stochastic gradient descent (SGD) on $K$ devices---where $E$ is a small constant and $K$ is a small fraction of the total devices in the network. The devices then communicate their model updates to a central server, where they are averaged.

While \fedavg has demonstrated empirical success in heterogeneous settings, it does not fully address the underlying challenges associated with heterogeneity.
In the context of systems heterogeneity, \fedavg does not allow participating devices to perform variable amounts of local work based on their underlying systems constraints; instead it is common to simply drop devices that fail to compute $E$ epochs within a specified time window~\citep{bonawitz2019towards}.  From a statistical perspective, \fedavg has been shown to diverge empirically in settings where the data is non-identically distributed across devices~\citep[e.g.,][Sec 3]{mcmahan2016FedAvg}. Unfortunately, \fedavg is difficult to analyze theoretically in such realistic scenarios and thus lacks convergence guarantees to characterize its behavior (see Section~\ref{sec:related_work} for additional details).

In this work, we propose \fedprox, a federated optimization algorithm that addresses the challenges of heterogeneity both theoretically and empirically. A key insight we have in developing \fedprox is that an interplay exists between systems and statistical heterogeneity in federated learning. Indeed, both dropping stragglers (as in \fedavg) or naively incorporating partial information from stragglers (as in \fedprox with the proximal term set to 0) implicitly increases statistical heterogeneity and can adversely impact convergence behavior. To mitigate this issue, we propose adding a proximal term to the objective that helps to improve the stability of the method. This term provides a principled way for the server to account for  heterogeneity associated with partial information. Theoretically, these modifications allow us to provide convergence guarantees for our method and to analyze the effect of heterogeneity. Empirically, we demonstrate that the modifications improve the stability and overall accuracy of federated learning in heterogeneous networks---improving the absolute testing accuracy by 22\% on average in highly heterogeneous settings. 

The remainder of this paper is organized as follows. In Section~\ref{sec:related_work}, we provide background on federated learning and an overview of related work. We then present our proposed framework, \fedprox, in Section~\ref{sec:methods}, and derive convergence guarantees for the framework accounting for both statistical and systems heterogeneity in Section~\ref{sec:conv_analysis}. Finally, in Section~\ref{sec:exps}, we provide a thorough empirical evaluation of \fedprox on a suite of synthetic and real-world federated datasets. Our empirical results help to illustrate and validate our theoretical analysis, and demonstrate the practical improvements of \fedprox over \fedavg in heterogeneous networks.

\section{Background and Related Work}\label{sec:related_work}
Large-scale machine learning, particularly in data center settings, has motivated the development of numerous distributed optimization methods in the past decade~\citep[see, e.g.,][]{Boyd:2010bw, Dekel:2012wm, Large_scale_DL_dean_2012, Zhang:2013wq,  param_server_Smola_14, shamir2014communication, AIDE_reddi_16,  elastic_SGD_zhang_LeCun_2015,  richtarik2016distributed, COCOA_Smith_2016}. 
However, as computing substrates such as phones, sensors, and wearable devices grow both in power and in popularity, it is increasingly attractive to learn statistical models locally in networks of distributed devices, in contrast to moving the data to the data center. This problem, known as federated learning, requires tackling novel challenges with privacy, heterogeneous data and devices, and massively distributed  networks~\cite{lireviewpaper}. 

Recent optimization methods have been proposed that are tailored to the specific challenges in the federated setting. These methods have shown significant improvements over traditional distributed approaches such as ADMM~\cite{Boyd:2010bw} or mini-batch methods~\cite{Dekel:2012wm} by allowing both for inexact local updating in order to balance communication vs. computation in large networks, and for a small subset of devices to be active at any communication round~\citep{mcmahan2016FedAvg, fed_multitask_smith_2017}. 
For example, \citet{fed_multitask_smith_2017} propose a communication-efficient primal-dual optimization method that learns separate but related models for each device through a multi-task learning framework. Despite the theoretical guarantees and practical efficiency of the proposed method, such an approach is not generalizable to non-convex problems, e.g., deep learning, where strong duality is no longer guaranteed.
In the non-convex setting, Federated Averaging (\fedavg), 
a heuristic method based on averaging local Stochastic Gradient Descent (SGD) updates in the primal, has instead been shown to work well empirically~\citep{mcmahan2016FedAvg}. 

Unfortunately, \fedavg is quite challenging to analyze due to its local updating scheme, the fact that few devices are active at each round, and the issue that data is frequently distributed in a heterogeneous nature in the network. In particular, as each device generates its own local data, \textit{statistical heterogeneity} is common with data being non-identically distributed between devices.
Several works have made steps towards analyzing \fedavg in simpler, non-federated settings. For instance, 
parallel SGD and related variants~\citep{elastic_SGD_zhang_LeCun_2015,shamir2014communication, AIDE_reddi_16, zhou2017convergence, local_SGD_stich_18, cooperative_SGD_Joshi_18,parallel_SGD_Srebro_18,lin2018don}, which make local updates similar to \fedavg, have been studied in the IID setting. However, the results rely on the premise that each local solver is a copy of the same stochastic process (due to the IID assumption). This line of reasoning does not apply to the heterogeneous setting. 

Although some recent works~\cite{yu2018parallel, wang2018adaptive, local_sgd_momentum_nonconvex, jiang2018linear} have explored convergence guarantees in statistically heterogeneous settings, they make the limiting assumption that all devices participate in each round of communication, which is often infeasible in realistic federated networks~\cite{mcmahan2016FedAvg}. Further, they rely on specific solvers to be used on each device (either SGD or GD), as compared to the solver-agnostic framework proposed herein, and add additional assumptions of convexity~\cite{wang2018adaptive} or uniformly bounded gradients~\cite{yu2018parallel} to their analyses.
There are also heuristic approaches that aim to tackle statistical heterogeneity by sharing the local device data or server-side proxy data~\citep{jeong2018communication, Zhao2018FederatedLW, huang2018loadaboost}. However, these methods may be unrealistic: in addition to imposing burdens on network bandwidth, sending local data to the server~\citep{jeong2018communication} violates the key privacy assumption of federated learning, and sending globally-shared proxy data to all devices~\citep{Zhao2018FederatedLW, huang2018loadaboost} requires effort to carefully generate or collect such auxiliary data. 

Beyond statistical heterogeneity, \emph{systems heterogeneity} is also a critical concern in federated networks. 
The storage, computational, and communication capabilities of
each device in federated networks may differ due to variability in hardware (CPU, memory), network
connectivity (3G, 4G, 5G, wifi), and power (battery level).
These system-level characteristics dramatically exacerbate challenges such as straggler mitigation and fault tolerance. One strategy used in practice is to ignore the more constrained devices failing to complete a certain amount of training~\cite{bonawitz2019towards}. However (as we demonstrate in Section~\ref{sec:exps}), this can have negative effects on convergence as it limits the number of effective devices contributing to training, and may induce bias in the device sampling procedure if the dropped devices have specific data characteristics.  

In this work, inspired by \fedavg, we explore a broader framework, \fedprox, that is capable of handling heterogeneous federated environments while maintaining similar privacy and computational benefits. We analyze the convergence behavior of the framework through a  \emph{statistical} dissimilarity characterization between local functions, while also taking into account practical \emph{systems} constraints. Our dissimilarity characterization is inspired by the randomized Kaczmarz method for solving linear system of equations \citep{linear_kaczmarz_1993,strohmer2009randomized}, a similar assumption of which has been used to analyze variants of SGD in other settings \citep[see, e.g.,][]{SGD_under_growth_schmidt_2013,vaswani2018fast,pmlr-v84-yin18a}.
Our proposed framework provides improved robustness and stability for optimization in heterogeneous federated networks.

Finally, in terms of related work, we note that two aspects of our proposed work---the proximal term in \fedprox and the bounded dissimilarity assumption used in our analysis---have been previously studied in the optimization literature, though often with very different motivations and in non-federated settings. For completeness, we provide a further discussion in Appendix~\ref{app: relwork} on this background work.

\section{Federated Optimization: Methods}\label{sec:methods}
In this section, we introduce the key ingredients behind recent methods for federated learning, including \fedavg, and then outline our proposed framework, \fedprox.  

Federated learning methods~\citep[e.g.,][]{mcmahan2016FedAvg, fed_multitask_smith_2017} are designed to handle multiple devices collecting data and a central server coordinating the global learning objective across the network. In particular, the aim is to minimize: 
\begin{align}
\label{eq:obj}
\min_w~f(w) = \sum_{k=1}^N p_k F_k(w)= \mathbb{E}_k [F_k(w)],
\end{align}
where $N$ is the number of devices, $p_k$ $\geq$ $0$, and $\sum_k p_k$=$1$. In general, the local objectives measure the local empirical risk over possibly differing data distributions $\mathcal{D}_k$, i.e., $F_k(w) \eqdef \E{x_k \sim \mathcal{D}_k}{f_k(w;x_k)}$, with $n_k$ samples available at each device $k$. Hence, we can set $p_k$=$\frac{n_k}{n}$, where $n$= $\sum_k n_k$ is the total number of data points. In this work, we consider $F_k(w)$ to be possibly non-convex.

To reduce communication, a common technique in federated optimization is that on each device, a \textit{local objective function} based on the device's data is used as a surrogate for the global objective function.
At each outer iteration, a subset of the devices are selected and \textit{local solvers} are used to optimize the local objective functions on each of the selected devices. The devices then communicate their {local model updates} to the central server, which aggregates them and updates the global model accordingly. The key to allowing flexible performance in this scenario is that each of the local objectives can be solved \textit{inexactly}. This allows the amount of local computation vs. communication to be tuned based on the number of local iterations that are performed (with additional local iterations corresponding to more exact local solutions). 
We introduce this notion formally below, as it will be utilized throughout the paper.
\begin{definition}[$\gamma$-inexact solution]\label{def: inexactness}
For a function $h(w; w_0) = F(w) + \frac{\mu}{2}\|w-w_0\|^2$, and $\gamma\in [0,1]$, we say $w^*$ is a $\gamma$-inexact solution of $\min_w h(w; w_0)$
if 
$\|\nabla h(w^*; w_0)\| \leq \gamma \|\nabla h(w_0; w_0)\|$, where $\nabla h(w; w_0) = \nabla F(w) + \mu (w-w_0)$.
Note that a smaller $\gamma$ corresponds to higher accuracy.
\end{definition}
We use $\gamma$-inexactness in our analysis (Section~\ref{sec:conv_analysis}) to measure the amount of local computation from the local solver at each round. {As discussed earlier, different devices are likely to make different progress towards solving the local subproblems due to variable systems conditions, and it is therefore important to allow $\gamma$ to vary both by device and by iteration. This is one of the motivations for our proposed framework discussed in the next sections. For ease of notation, we first derive our main convergence results  assuming a uniform $\gamma$ as defined here (Section \ref{sec:conv_analysis}), and then provide results with variable $\gamma$'s in Corollary \ref{coro: variable_gamma}.} 

\subsection{Federated Averaging~(\fedavg)}\label{sec:fedavg}

In Federated Averaging (\fedavg)~\citep{mcmahan2016FedAvg}, the local surrogate of the global objective function at device $k$ is  $F_k\left(\cdot\right)$, and the local solver is stochastic gradient descent~(SGD), with the same learning rate and number of local epochs used on each device. At each round, a subset $K \ll N$ of the total devices are selected and run SGD locally for $E$ number of epochs, and then the resulting model updates are averaged. The details of \fedavg are summarized in Algorithm \ref{alg:FEDAVG}. 

\setlength{\textfloatsep}{10pt}
\begin{algorithm}[h]
    \begin{algorithmic}
	\caption{Federated Averaging (\fedavg)}
	\label{alg:FEDAVG}
	\STATE {\bf Input:}  $K$, $T$, $\eta$, $E$, $w^0$, $N$, $p_k$, $k=1,\cdots, N$
	\FOR  {$t=0, \cdots, T-1$}
		\STATE Server selects a subset $S_t$ of $K$ devices at random (each device $k$ is chosen with probability $p_k$)
		\STATE Server sends $w^t$ to all chosen devices 
		\STATE Each device $k \in S_t$ updates $w^t$ for $E$ epochs of SGD on $F_k$ with step-size $\eta$ to obtain $w_k^{t+1}$
		\STATE Each device $k \in S_t$ sends $w_k^{t+1}$ back to the server
		\STATE Server aggregates the $w$'s as {\small$w^{t+1} = \frac{1}{K}\sum_{k \in S_t} w_k^{t+1}$}
	\ENDFOR
	\end{algorithmic}
\end{algorithm}

\citet{mcmahan2016FedAvg} show empirically that it is crucial to tune the optimization hyperparameters of \fedavg properly. In particular, the number of local epochs in \fedavg plays an important role in convergence. On one hand, performing more local epochs allows for more local computation and potentially reduced communication, which can greatly improve the overall convergence speed in communication-constrained networks.  On the other hand, 
with dissimilar (heterogeneous) local objectives $F_k$, a larger number of local epochs may lead each device towards the optima of its local objective as opposed to the global objective---potentially hurting convergence or even causing the method to diverge. Further, in federated networks with heterogeneous systems resources, setting the number of local epochs to be high may increase the risk that devices do not complete training within a given communication round and must therefore drop out of the procedure~\cite{bonawitz2019towards}.

In practice, it is therefore important to find a way to set the local epochs to be high (to reduce communication) while also allowing for robust convergence.  More fundamentally, we note that the `best' setting for the number of local epochs is likely to change at each iteration and on each device---as a function of both the local data and available systems resources. 
Indeed, a more natural approach than mandating a \textit{fixed} number of local epochs is to allow the epochs to \textit{vary} according to the characteristics of the network, and to carefully merge solutions by accounting for this heterogeneity.  We formalize this strategy in \fedprox, introduced below.

\subsection{Proposed Framework: \fedprox}\label{sec:fedprox}

Our proposed framework, \fedprox (Algorithm~\ref{alg:FEDPROX}), is similar to \fedavg in that a subset of devices are selected at each round, local updates are performed, and these updates are then averaged to form a global update. {However, \fedprox makes the following simple yet critical modifications, which result in significant empirical improvements and also allow us to provide convergence guarantees for the method.}

\textbf{Tolerating partial work.} As previously discussed, different devices in federated networks often have different resource constraints in terms of the computing hardware, network connections, and battery levels. Therefore, it is unrealistic to force each device to perform a uniform amount of work (i.e., running the same number of local epochs, $E$), as in \fedavg. 
In \fedprox, we generalize \fedavg by allowing for variable amounts of work to be performed locally across devices based on their available systems resources, and then aggregate the partial solutions sent from the stragglers (as compared to dropping these devices). In other words, instead of assuming a uniform $\gamma$ for all devices throughout the training process, \fedprox implicitly accommodates variable $\gamma$'s for different devices and at different iterations. We formally define $\gamma_k^t$-inexactness for device $k$ at iteration $t$ below, which is a natural extension from Definition \ref{def: inexactness}.

\begin{definition}[$\gamma_k^t$-inexact solution]\label{def: variable_inexactness}
For a function $h_k(w; w_t) = F_k(w) + \frac{\mu}{2}\|w-w_t\|^2$, and $\gamma\in [0,1]$, we say $w^*$ is a $\gamma_k^t$-inexact solution of $\min_w h_k(w; w_t)$
if $\|\nabla h_k(w^*; w_t)\| \leq \gamma_k^t \|\nabla h_k(w_t; w_t)\|$, where $\nabla h_k(w; w_t) = \nabla F_k(w) + \mu (w-w_t)$.
Note that a smaller $\gamma_k^t$ corresponds to higher accuracy.
\end{definition}

{Analogous to Definition~\ref{def: inexactness}, $\gamma_k^t$ measures how much local computation is performed to solve the local subproblem on device $k$ at the $t$-th round. The variable number of local iterations can be viewed as a proxy of $\gamma_k^t$. Utilizing the more flexible $\gamma_k^t$-inexactness, we can readily extend the convergence results under Definition \ref{def: inexactness}  (Theorem \ref{lemma: decrease_0}) to consider issues related to systems heterogeneity such as stragglers (see Corollary \ref{coro: variable_gamma}).}

\textbf{Proximal term.} As mentioned in Section \ref{sec:fedavg}, while tolerating nonuniform amounts of work to be performed across devices can help alleviate negative impacts of systems heterogeneity, too many local updates may still (potentially) cause the methods to diverge due to the underlying heterogeneous data. We propose to add a proximal term to the local subproblem to effectively limit the impact of variable local updates. In particular, instead of just minimizing the local function $F_k(\cdot)$, device $k$ uses its local solver of choice to approximately minimize the following objective $h_k$:
\begin{align}
\min_w 
h_k(w; ~w^{t}) = F_k(w) + \frac{\mu}{2}\|w-w^{t}\|^2 \, .
\end{align}

The proximal term is beneficial in two aspects: (1) It addresses the issue of statistical heterogeneity by restricting the local updates to be closer to the initial (global) model without any need to manually set the number of local epochs. (2) It allows for safely incorporating variable amounts of local work resulting from systems heterogeneity. We summarize the steps of \fedprox in Algorithm \ref{alg:FEDPROX}.

\begin{algorithm}[h]
	\begin{algorithmic}
		\caption{\fedprox (Proposed Framework)}
		\label{alg:FEDPROX}	
		\STATE {\bf Input:}  $K$, $T$, $\mu$, $\gamma$, $w^0$, $N$, $p_k$, $k=1,\cdots, N$
		\FOR  {$t=0, \cdots, T-1$}
			\STATE Server selects a subset $S_t$ of $K$ devices at random (each device $k$ is chosen with probability $p_k$)
			\STATE Server sends $w^t$ to all chosen devices
			\STATE Each chosen device $k \in S_t$ finds a $w_k^{t+1}$ which is a $\gamma_k^t$-inexact minimizer of:
			$w_k^{t+1} \approx \arg\min_w~h_k(w; ~w^{t}) = F_k(w) + \frac{\mu}{2}\|w-w^{t}\|^2$
			\STATE Each device $k \in S_t$ sends $w_k^{t+1}$ back to the server
			\STATE Server aggregates the $w$'s as
			{\small$w^{t+1} = \frac{1}{K} \sum_{k \in S_t} w_k^{t+1}$}
		\ENDFOR
	\end{algorithmic}
\end{algorithm}

We note that proximal terms such as the one above are a popular tool utilized throughout the optimization literature; for completeness, we provide a more detailed discussion on this in Appendix~\ref{app: relwork}. An important distinction of the proposed usage is that we suggest, explore, and analyze such a term for the purpose of tackling heterogeneity in federated networks. Our analysis (Section~\ref{sec:conv_analysis}) is also unique in considering solving such an objective in a distributed setting with: (1) non-IID partitioned data, (2) the use of any local solver, (3) variable inexact updates across devices, and (4) a subset of devices being active at each round. These assumptions are critical to providing a characterization of such a framework in realistic federated scenarios.

In our experiments (Section~\ref{sec:exps}), we demonstrate that tolerating partial work is beneficial in the presence of systems heterogeneity and our modified local subproblem in \fedprox results in more robust and stable convergence compared to vanilla \fedavg for heterogeneous datasets. In Section~\ref{sec:conv_analysis}, we also see that the usage of the proximal term makes \fedprox more amenable to theoretical analysis (i.e., the local objective may be more well-behaved). In particular, if $\mu$ is chosen accordingly, the Hessian of $h_k$ may be positive semi-definite. Hence, when $F_k$ is non-convex, $h_k$ will be convex, and when $F_k$ is convex, it becomes $\mu$-strongly convex. 

{Finally, we note that since \fedprox makes only lightweight modifications to \fedavg, this allows us to reason about the behavior of the widely-used \fedavg method, and enables easy integration of \fedprox into existing packages/systems, such as TensorFlow Federated and LEAF~\cite{TFF, caldas2018leaf}. In particular, we note that \fedavg is a special case of \fedprox with (1) $\mu=0$, (2) the local solver specifically chosen to be SGD, and (3) a constant $\gamma$ (corresponding to the number of local epochs) across devices and updating rounds (i.e., no notion of systems heterogeneity). \fedprox is in fact much more general in this regard, as it allows for partial work to be performed across devices and any local (possibly non-iterative) solver to be used on each device.}

\section{\fedprox: Convergence Analysis}
\label{sec:conv_analysis}

\fedavg and \fedprox are stochastic algorithms by nature: in each round, only a fraction of the devices are sampled to perform the update, and the updates performed on each device may be inexact. It is well known that in order for stochastic methods to converge to a stationary point, a decreasing step-size is required. This is in contrast to non-stochastic methods, e.g., gradient descent, that can find a stationary point by employing a constant step-size. In order to analyze the convergence behavior of methods with constant step-size (as is usually implemented in practice), we need to quantify the degree of dissimilarity among the local objective functions. This could be achieved by assuming the data to be IID, i.e., homogeneous across devices. Unfortunately, in realistic federated networks, this assumption is impractical. Thus, we first propose a metric that specifically measures the dissimilarity among local functions (Section~\ref{sec:dissimilarity}), and then analyze \fedprox under this assumption while allowing for variable $\gamma$'s (Section~\ref{sec:fedproxconv}).

\subsection{Local dissimilarity}\label{sec:dissimilarity}

Here we introduce a measure of dissimilarity between the devices in a federated network, which is sufficient to prove convergence. This can also be satisfied via a simpler and more restrictive bounded variance assumption of the gradients (Corollary \ref{coro: bounded_var_sim}), which we explore in our experiments in Section~\ref{sec:exps}. Interestingly, similar assumptions~\citep[e.g.,][]{SGD_under_growth_schmidt_2013,vaswani2018fast, pmlr-v84-yin18a} have been explored elsewhere but for differing purposes; we provide a  discussion of these works in Appendix~\ref{app: relwork}.

\begin{definition}[$B$-local dissimilarity]\label{def: similarity}
The local functions $F_k$ are $B$-locally dissimilar at $w$ if ${\E{k}{\|\nabla F_k(w)\|^2}}\! \leq\!{\|\nabla f(w)\|^2}B^2$. We further define $B(w)\!=\! \sqrt{\frac{\E{k}{\|\nabla F_k(w)\|^2}}{\|\nabla f(w)\|^2}}$ for\footnote{As an exception we define $B(w)=1$ when $\E{k}{\|\nabla F_k(w)\|^2}=\|\nabla f(w)\|^2$, i.e. $w$ is a stationary solution that all the local functions $F_k$ agree on.} $\|\nabla f(w)\|\!\neq\!0$. 
\end{definition}

Here $\E{k}{\cdot}$ denotes the expectation over devices with masses $p_{k}=n_{k}/n$ and $\sum_{k=1}^{N}p_{k}=1$ (as in Equation~\ref{eq:obj}).
Definition \ref{def: similarity} can be seen as a generalization of the IID assumption  with  bounded  dissimilarity, while allowing for statistical heterogeneity. 
As a sanity check, when all the local functions are the same, we have $B(w) = 1$ for all $w$.
However, in the federated setting, the data distributions are often heterogeneous and $B >1$ due to sampling discrepancies even if the samples are assumed to be IID. Let us also consider the case where $F_k\left(\cdot\right)$'s are associated with empirical risk objectives. If the samples on all the devices are homogeneous, i.e., they are sampled in an IID fashion, then as $\min_k~n_k$ $\to \infty$, it follows that $B(w) \to 1$ for every $w$ as all the local functions converge to the same expected risk function in the large sample limit.  Thus, $B(w) \geq 1$ and the larger the value of $B(w)$, the larger is the dissimilarity among the local functions.

Using Definition~\ref{def: similarity}, we now state our formal dissimilarity assumption, which we use in our convergence analysis. This simply requires that the dissimilarity defined in Definition~\ref{def: similarity} is bounded.  As discussed later, our convergence rate is a function of the statistical heterogeneity/device dissimilarity in the network.

\begin{assumption}[Bounded dissimilarity]\label{assumption: 1}
For some $\epsilon>0$, there exists a $B_\epsilon$ such that for all the points $w\in \mathcal{S}_\epsilon^c = \{w~|~\|\nabla f(w)\|^{2}> \epsilon\}$, $B(w)\leq B_\epsilon$.
\end{assumption}
For most practical machine learning problems, there is no need to solve the problem to highly accurate stationary solutions, i.e., $\epsilon$ is typically not very small. Indeed, it is well-known that solving the problem beyond some threshold may even hurt generalization performance due to overfitting~\cite{yao2007early}. Although in practical federated learning problems the samples are not IID, they are still sampled from distributions that are not entirely unrelated (if this were the case, e.g., fitting a single global model $w$ across devices would be ill-advised). Thus, it is reasonable to assume that the dissimilarity between local functions remains bounded throughout the training process. We also measure the dissimilarity metric empirically on real and synthetic datasets in Section~\ref{section: sim} and show that this metric captures real-world statistical heterogeneity and is correlated with practical performance (the smaller the dissimilarity, the better the convergence). 

\vspace{.25em}
\subsection{\fedprox Analysis}
\label{sec:fedproxconv}
Using the bounded dissimilarity assumption (Assumption~\ref{assumption: 1}), we now analyze the amount of expected decrease in the objective when one step of \fedprox is performed. Our convergence rate (Theorem \ref{coro: conv_FEDPROX}) can be directly derived from the results of the expected decrease per updating round. We assume the same $\gamma_k^t$ for any $k,t$ for ease of notation in the following analyses. 

\vspace{.5em}
\begin{theorem}[Non-convex \fedprox convergence: $B$-local dissimilarity]\label{lemma: decrease_0}
Let Assumption~\ref{assumption: 1} hold.
Assume the functions $F_k$ are non-convex, $L$-Lipschitz smooth, and there exists $L_- > 0$, such that $\nabla^2 F_k\succeq -L_- \mathbf{I}$, with $\bar{\mu} \eqdef \mu-L_->0$. Suppose that $w^t$ is not a stationary solution and the local functions $F_k$ are $B$-dissimilar, i.e. $B(w^t)\leq B$. If $\mu$, $K$, and $\gamma$ in Algorithm \ref{alg:FEDPROX} are chosen such that
\medmuskip=1mu
\thinmuskip=1mu
\thickmuskip=1mu
\begin{align*}
&\rho = \left(\frac{1}{\mu} - \frac{\gamma B}{\mu}-\frac{B(1+\gamma)\sqrt{2}}{\bar{\mu}\sqrt{K}} - \frac{LB(1+\gamma)}{\bar{\mu}\mu} \right.\nonumber \\& \left.- \frac{L(1+\gamma)^2B^2}{2\bar{\mu}^2} - \frac{LB^2(1+\gamma)^2}{\bar{\mu}^2 K}\bigg({2\sqrt{2K}+2}\bigg) \right)>0 ,
\end{align*}
then at iteration $t$ of Algorithm \ref{alg:FEDPROX}, we have the following expected decrease in the global objective: $$\E{S_t}{f(w^{t+1})} \leq f(w^{t}) - \rho \|\nabla f(w^t)\|^2,$$ where $S_t$ is the set of $K$ devices chosen at iteration $t$.
\end{theorem}

We direct the reader to Appendix \ref{sec:main_res_proof} for a detailed proof. The key steps include applying our notion of $\gamma$-inexactness (Definition \ref{def: inexactness}) for each subproblem and using the bounded dissimilarity assumption, while allowing for  only $K$ devices to be active at each round. This last step in particular introduces $\mathbb{E}_{S_t}$, an expectation with respect to the choice of devices, $S_t$, in round $t$. We note that in our theory, we require $\bar{\mu} > 0$, which is a sufficient but not necessary condition for \fedprox to converge. Hence, it is possible that some other $\mu$ (not necessarily satisfying $\bar{\mu} > 0$) can also enable convergence, as we explore empirically (Section \ref{sec:exps}).

Theorem \ref{lemma: decrease_0} uses the dissimilarity in Definition~\ref{def: similarity} to identify sufficient decrease of the objective value at each iteration for \fedprox. In Appendix \ref{app: boundedvar}, we provide a corollary characterizing the performance with a more common (though slightly more restrictive) bounded variance assumption. This assumption is commonly employed, e.g., when analyzing methods such as SGD.  
We next provide sufficient (but not necessary) conditions that ensure $\rho > 0$ in Theorem \ref{lemma: decrease_0} such that sufficient decrease is attainable after each round.

\vspace{.5em}
\begin{remark}
For $\rho$ in Theorem \ref{lemma: decrease_0} to be positive, we need $\gamma B<1$ and $\tfrac{B}{\sqrt{K}}<1$.
These conditions help to quantify the trade-off between dissimilarity (B) and the algorithm parameters ($\gamma$, $K$). 
\end{remark}

Finally, we can use the above sufficient decrease to the characterize the rate of convergence to the set of approximate stationary solutions $\mathcal{S}_s = \{w~|~\E{}{\|\nabla f(w)\|^2}\leq \epsilon\}$ under the bounded dissimilarity assumption, Assumption~\ref{assumption: 1}. Note that these results hold for general non-convex $F_k(\cdot)$.

\vspace{.5em}
\begin{theorem}[Convergence rate: \fedprox]\label{coro: conv_FEDPROX}
Given some $\epsilon>0$, assume that for $B\geq B_\epsilon$, $\mu$, $\gamma$, and $K$ the assumptions of Theorem \ref{lemma: decrease_0} hold at each iteration of \fedprox. Moreover, $f(w^0)-f^* = \Delta$. Then, after $T = O(\frac{\Delta}{\rho \epsilon})$ iterations of \fedprox, we have $\frac{1}{T}\sum_{t=0}^{T-1}\E{}{\|\nabla f(w^t)\|^2} \leq \epsilon$.
\end{theorem}

While the results thus far hold for non-convex $F_k(\cdot)$, we can also characterize the convergence for the special case of convex loss functions with exact minimization in terms of local objectives (Corollary \ref{coro: convex_exact_fedprox}). A proof is provided in Appendix~\ref{app: convexfedprox}.

\vspace{.5em}
\begin{corollary}[Convergence: Convex case]
\label{coro: convex_exact_fedprox}
Let the assertions of Theorem \ref{lemma: decrease_0} hold. In addition, let $F_k\left(\cdot\right)$'s be convex and $\gamma_k^t=0$ for any $k,t$, i.e., all the local problems are solved exactly, if $1\ll B\leq 0.5\sqrt{K}$, then we can choose ${\mu}\approx {6LB^2}$ from which it follows that $\rho\approx \frac{1}{24LB^2}$.
\end{corollary}

Note that small $\epsilon$ in Assumption~\ref{assumption: 1} translates to larger $B_\epsilon$. Corollary \ref{coro: convex_exact_fedprox} suggests that, in order to solve the problem with increasingly higher accuracies using \fedprox, one needs to increase $\mu$ appropriately. We empirically verify that $\mu>0$ leads to more stable convergence in Section \ref{sec: exp_statistical}. Moreover, in Corollary \ref{coro: convex_exact_fedprox}, if we plug in the upper bound for $B_\epsilon$, under a bounded variance assumption (Corollary \ref{coro: bounded_var_sim}), the number of required steps to achieve accuracy $\epsilon$ is $O(\frac{L\Delta}{\epsilon} + \frac{L\Delta\sigma^2}{\epsilon^2})$.
Our analysis helps to characterize the performance of \fedprox and similar methods when local functions are dissimilar.

\begin{remark}[Comparison with SGD]
\label{rem:sgd}
We note that \fedprox achieves the same asymptotic convergence guarantee as SGD: Under the bounded variance assumption, for small $\epsilon$, if we replace $B_\epsilon$ with its upper-bound in Corollary \ref{coro: bounded_var_sim} and choose $\mu$ large enough, the iteration complexity of \fedprox when the subproblems are solved exactly and $F_k(\cdot)$'s are convex is $O(\frac{L\Delta}{\epsilon} + \frac{L\Delta\sigma^2}{\epsilon^2})$, the same as SGD \cite{ghadimi2013stochastic}.
\end{remark}

To provide context for the rate in Theorem~\ref{coro: conv_FEDPROX}, we compare it with SGD in the convex case in Remark~\ref{rem:sgd}. 
In general, our analysis of \fedprox does not yield convergence rates that improve upon classical distributed SGD (without local updating)---even though \fedprox possibly performs more work locally at each communication round. In fact, when data are generated in a non-identically distributed fashion, it is possible for local updating schemes such as \fedprox to perform worse than distributed SGD. Therefore, our theoretical results do not necessarily demonstrate the superiority of \fedprox over distributed SGD; rather, they provide sufficient (but not necessary) conditions for \fedprox to converge. Our analysis is the first we are aware of to analyze any federated (i.e., with local-updating schemes and low device participation) optimization method for Problem~\eqref{eq:obj} in heterogeneous settings.

Finally, we note that the previous analyses assume no systems heterogeneity and use the same $\gamma$ for all devices and iterations. However, we can extend them to allow for $\gamma$ to vary by device and by iteration (as in Definition \ref{def: variable_inexactness}), which corresponds to allowing devices to perform variable amounts of work as determined by the local systems conditions. We provide convergence results with variable $\gamma$'s below. 
\begin{corollary}[Convergence: Variable $\gamma$'s]
\label{coro: variable_gamma}
Assume the functions $F_k$ are non-convex, $L$-Lipschitz smooth, and there exists $L_- > 0$, such that $\nabla^2 F_k\succeq -L_- \mathbf{I}$, with $\bar{\mu} \eqdef \mu-L_->0$. Suppose that $w^t$ is not a stationary solution and the local functions $F_k$ are $B$-dissimilar, i.e. $B(w^t)\leq B$. If $\mu$, $K$, and $\gamma_k^t$ in Algorithm \ref{alg:FEDPROX} are chosen such that
\medmuskip=1mu
\thinmuskip=1mu
\thickmuskip=1mu
\begin{align*}
&\rho^t = \left(\frac{1}{\mu} - \frac{\gamma^t B}{\mu}-\frac{B(1+\gamma^t)\sqrt{2}}{\bar{\mu}\sqrt{K}} - \frac{LB(1+\gamma^t)}{\bar{\mu}\mu} \right.\nonumber \\& \left.- \frac{L(1+\gamma^t)^2B^2}{2\bar{\mu}^2} - \frac{LB^2(1+\gamma^t)^2}{\bar{\mu}^2 K}\bigg({2\sqrt{2K}+2}\bigg) \right)>0 ,
\end{align*}
then at iteration $t$ of Algorithm \ref{alg:FEDPROX}, we have the following expected decrease in the global objective: $$\E{S_t}{f(w^{t+1})} \leq f(w^{t}) - \rho^t \|\nabla f(w^t)\|^2,$$ where $S_t$ is the set of $K$ devices chosen at iteration $t$ and $\gamma_t=\max_{k \in S_t}~{\gamma_k^t}$.
\end{corollary}

The proof can be easily extended from the proof for Theorem \ref{lemma: decrease_0} , noting the fact that $\mathbb{E}_k[(1+\gamma_k^t) \|\nabla F_k(w^t)\|] \leq (1+\max_{k \in S_t}~\gamma_k^t) \mathbb{E}_k[\|\nabla F_k(w^t)\|]$.

\section{Experiments}
\label{sec:exps}
We now present empirical results for the generalized \fedprox framework. In Section \ref{sec: exp_systems}, we demonstrate the improved performance of \fedprox tolerating partial solutions in the face of systems heterogeneity. In Section \ref{sec: exp_statistical}, we show the effectiveness of \fedprox in the settings with statistical heterogeneity (regardless of systems heterogeneity). We also study the effects of statistical heterogeneity on convergence (Section \ref{sec: exp_synthetic_statistical}) and show how empirical convergence is related to our theoretical bounded dissimilarity assumption (Assumption~\ref{assumption: 1}) (Section \ref{sec: exp_dissimilar}). We provide thorough details of the experimental setup in Section \ref{section: set up} and Appendix~\ref{app:exps}. All code, data, and experiments are publicly available at: \href{https://github.com/litian96/FedProx}{github.com/litian96/FedProx}. 

\subsection{Experimental Details}\label{section: set up}

We evaluate \fedprox on diverse tasks, models, and real-world federated datasets. In order to better characterize statistical heterogeneity and study its effect on convergence, we also evaluate on a set of synthetic data, which allows for more precise manipulation of statistical heterogeneity. We simulate systems heterogeneity by assigning different amounts of local work to different devices.

\textbf{Synthetic data.} To generate synthetic data, we follow a similar setup to that  in~\citet{shamir2014communication}, additionally imposing heterogeneity among devices. In particular, for each device $k$, we generate samples $(X_k, Y_k)$ according to the model $y=\textrm{\emph{argmax}(softmax}(Wx+b))$, $x \in \mathbb{R}^{60}, W \in \mathbb{R}^{10 \times 60}, b \in \mathbb{R}^{10}$. 
We model $W_k \sim \mathcal{N}(u_k, 1)$, 
$b_k \sim \mathcal{N}(u_k, 1)$, $u_k \sim \mathcal{N}(0, \alpha)$; $x_k \sim \mathcal{N}(v_k, \Sigma)$, where the covariance matrix $\Sigma$ is diagonal with $\Sigma_{j,j}=j^{-1.2}$. Each element in the mean vector $v_k$ is drawn from $\mathcal{N}(B_k, 1), B_{k} \sim N(0, \beta)$. Therefore, $\alpha$ controls how much local models differ from each other and $\beta$ controls how much the local data at each device differs from that of other devices. We vary $\alpha, \beta$ to generate three heterogeneous distributed datasets, denoted Synthetic ($\alpha, \beta$), as shown in Figure \ref{fig: synthetic_statistical}. We also generate one IID dataset by setting the same $W, b$ 
on all devices and setting $X_k$ to follow the same distribution. 
Our goal is to learn a global $W$ and $b$. Full details are given in Appendix \ref{app:data}.

\textbf{Real data.} We also explore four real datasets;  statistics are 
summarized in Table~\ref{table: data}. These datasets are curated from prior work in federated learning as well as recent federated learning benchmarks~\cite{mcmahan2016FedAvg,caldas2018leaf}. 
We study a convex classification problem with MNIST~\cite{lecun1998gradient} using multinomial logistic regression. To impose statistical heterogeneity, we distribute the  data  among 1,000 devices such that each device has samples of only two digits and the number of samples per device follows a power law. We then study a more complex 62-class Federated Extended  MNIST~\cite{cohen2017emnist,caldas2018leaf} (FEMNIST)  dataset using the same model. For the non-convex setting, we consider a text sentiment analysis task on tweets from Sentiment140 (Go et al., 2009) (Sent140) with an LSTM classifier, where each twitter account corresponds to a device. We also investigate the task of next-character prediction on the dataset of \emph{The Complete Works of William Shakespeare}~\cite{mcmahan2016FedAvg} (Shakespeare). Each speaking role in the plays is associated with a different device.
Details of datasets, models, and workloads are provided in Appendix~\ref{app:data}. 

\begin{table}[h]
	\begin{center}
	    \vspace{-2mm}
		\caption{Statistics of four real federated datasets.}
		\vspace{0.15in}
		\label{table: data}
		\begin{tabular}{ lllll } 
			\toprule
			\textbf{Dataset} & \textbf{Devices} & \textbf{Samples} & 
			\multicolumn{2}{l}{\textbf{Samples/device}} \\
			\cmidrule(l){4-5}
			& &  & mean & stdev \\
			\hline
			MNIST & 1,000 & 69,035 & 69 & 106  \\
			FEMNIST  & 200 & 18,345 & 92 & 159 \\
			Shakespeare & 143 & 517,106 & 3,616 & 6,808 \\
			Sent140 & 772 &  40,783 & 53 & 32 \\
			\bottomrule
		\end{tabular}
	\end{center}
	\vspace{-2mm}
\end{table}

\textbf{Implementation.} We implement \fedavg (Algorithm \ref{alg:FEDAVG}) and \fedprox (Algorithm \ref{alg:FEDPROX}) in Tensorflow \cite{tensorflow2015-whitepaper}. In order to draw a fair comparison with \fedavg, we employ SGD as a local solver for \fedprox, and adopt a slightly different device sampling scheme than that in Algorithms~\ref{alg:FEDAVG} and~\ref{alg:FEDPROX}: sampling devices uniformly and then averaging the updates with weights proportional to the number of local data points (as originally proposed in~\citet{mcmahan2016FedAvg}). While this sampling scheme is not supported by our analysis, we observe similar relative behavior of \fedprox vs. \fedavg whether or not it is employed. Interestingly, we also observe that the sampling scheme proposed herein in fact results in more stable performance for both methods (see Appendix~\ref{app:compare_scheme}, Figure~\ref{fig: compare_sampling}). This suggests an additional benefit of the proposed framework. Full details are provided in Appendix \ref{app:implementation}.

\textbf{Hyperparameters \& evaluation metrics.} For each dataset, we tune the learning rate on \fedavg (with $E$=1 and without systems heterogeneity) and use the same learning rate for all experiments on that dataset. We set the number of selected devices to be 10 for all experiments on all datasets. 
For each comparison, we fix the randomly selected devices, the stragglers, and mini-batch orders across all runs. We report all metrics based on the global objective $f(w)$. Note that in our simulations (see Section \ref{sec: exp_systems} for details), we assume that each communication round corresponds to a specific aggregation time stamp (measured in real-world global wall-clock time)---we therefore report results in terms of rounds rather than FLOPs or wall-clock time. 
See details of the hyper-parameters in Appendix~\ref{app:implementation}.

\subsection{Systems Heterogeneity:  Tolerating Partial Work} \label{sec: exp_systems}
In order to measure the effect of allowing for partial solutions to be sent to handle systems heterogeneity with \fedprox, we simulate federated settings with varying system heterogeneity, as described below.

\textbf{Systems heterogeneity simulations.}
We assume that there exists a global clock during training, and each participating device determines the amount of local work as a function of this clock cycle and its systems constraints. This specified amount of local computation corresponds to some implicit
value $\gamma_k^t$ for device $k$ at the $t$-th iteration. In our simulations, we fix a global number of epochs $E$, and force some devices to perform fewer updates than $E$ epochs given their current systems constraints. In particular, for varying heterogeneous settings, at
each round, we assign $x$ number of epochs (chosen uniformly at random between [1, $E$]) to 0\%, 50\%, and 90\% of the selected devices, respectively. Settings where 0\% devices perform fewer than $E$ epochs of work correspond to the environments \textit{without} systems heterogeneity, while 90\% of the devices sending their partial solutions corresponds to highly heterogeneous environments.  \fedavg will simply drop these 0\%, 50\%, and 90\% stragglers upon reaching the global clock cycle, and \fedprox will incorporate the partial updates from these devices.

\begin{figure*}[t]
    \centering
    \includegraphics[width=\textwidth, trim={0mm 0mm 0mm 4mm},clip]{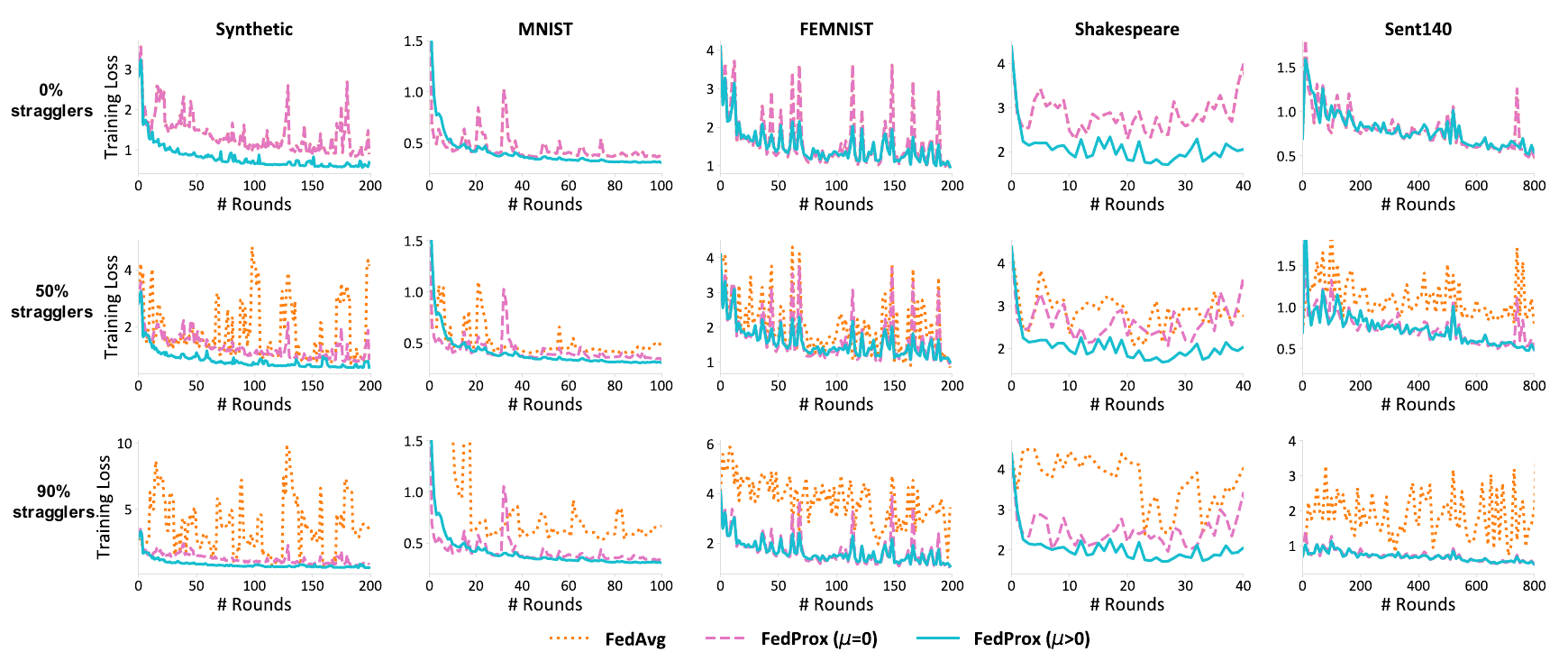}
    \vspace{-7mm}
    \caption{\fedprox results in significant convergence improvements relative to \fedavg in heterogeneous networks. We simulate different levels of systems heterogeneity by forcing 0\%, 50\%, and 90\% devices to be the stragglers (dropped by \fedavg). (1) Comparing \fedavg and \fedprox ($\mu=0$), we see that allowing for variable amounts of work to be performed can help convergence in the presence of systems heterogeneity. (2) Comparing \fedprox ($\mu=0$) with \fedprox ($\mu>0$), we show the benefits of our added proximal term. \fedprox with $\mu>0$ leads to more stable convergence and enables otherwise divergent methods to converge, both in the presence of systems heterogeneity (50\% and 90\% stragglers) and without systems heterogeneity (0\% stragglers). Note that \fedprox with $\mu=0$ and without systems heterogeneity (no stragglers) corresponds to \fedavg. 
    We also report testing accuracy in Figure \ref{fig:accuracy_full_e20}, Appendix \ref{appen:complete_result}, and show that \fedprox improves the test accuracy on all datasets.}
    \vspace{-3mm}
    \label{fig:loss_all_e20}
\end{figure*}

\begin{figure*}
    \centering
    \begin{subfigure}{1\textwidth}
        \centering
        \includegraphics[width=0.9\textwidth,trim={3mm 3mm 3mm 7mm},clip]{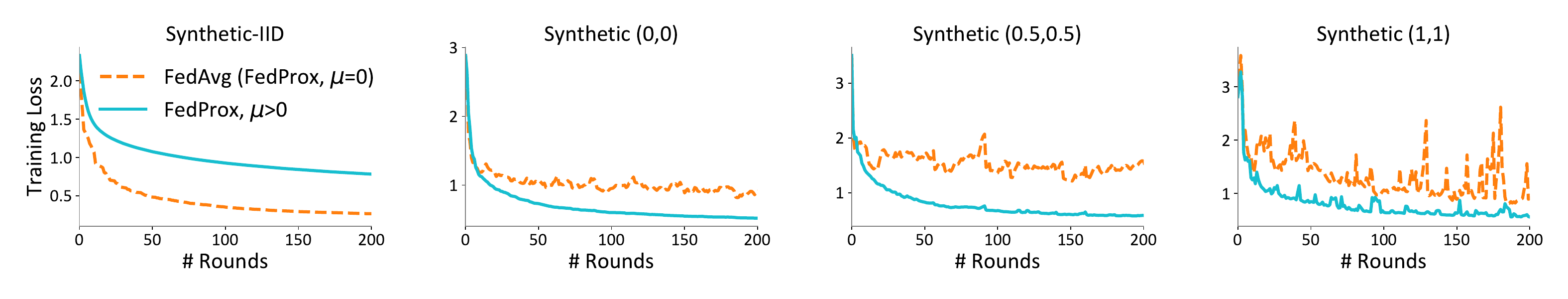}
    \end{subfigure}
    \begin{subfigure}{1\textwidth}
        \centering
        \includegraphics[width=0.9\textwidth,trim={3mm 7mm 3mm 4mm},clip]{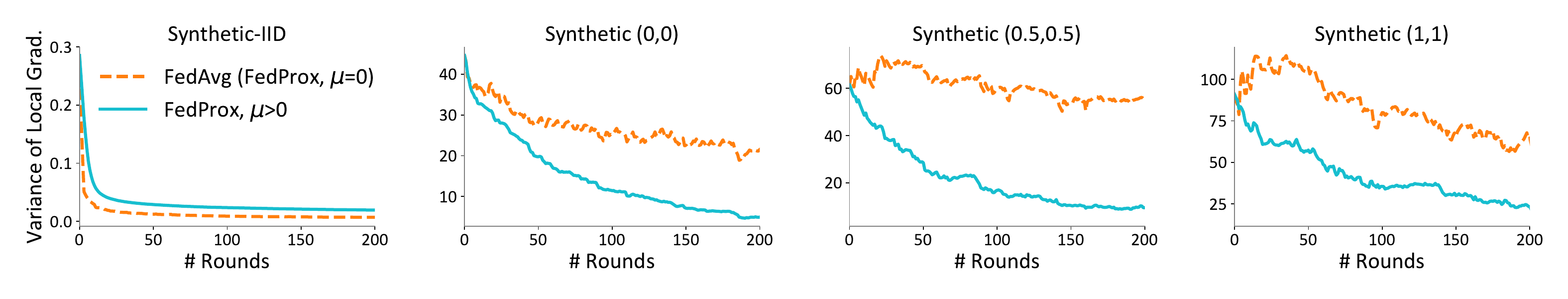}
    \end{subfigure}
    \vspace{-4mm}
    \caption{Effect of data heterogeneity on convergence. We remove the effects of systems heterogeneity by forcing each device to run the same amount of epochs. In this setting, \fedprox with $\mu=0$ reduces to \fedavg. (1) Top row: We show training loss  (see results on testing accuracy in Appendix~\ref{app:full experiments}, Figure~\ref{fig: non-iidness_all}) on four synthetic datasets whose statistical heterogeneity increases from left to right. Note that the method with $\mu=0$ corresponds to \fedavg. Increasing heterogeneity leads to worse convergence, but setting $\mu > 0$ can help to combat this. (2) Bottom row: We show the corresponding dissimilarity measurement (variance of gradients) of the four synthetic datasets. This metric captures statistical heterogeneity and is consistent with training loss --- smaller dissimilarity indicates better convergence.}
    \vspace{-2mm}
    \label{fig: synthetic_statistical}
\end{figure*}

In Figure \ref{fig:loss_all_e20}, we set $E$ to be 20 and study the effects of aggregating partial work from the otherwise dropped devices. The synthetic dataset here is taken from Synthetic (1,1) in Figure \ref{fig: synthetic_statistical}. We see that on all the datasets, systems heterogeneity has negative effects on convergence, and larger heterogeneity results in worse convergence (\fedavg). Compared with dropping the more constrained devices (\fedavg), incorporating variable amounts of work (\fedprox, $\mu=0$) is beneficial and leads to more stable and faster convergence. We also observe that setting $\mu > 0$ in \fedprox can further improve convergence, as we discuss in Section~\ref{sec: exp_statistical}.

We additionally investigate two less heterogeneous settings. First, we limit the capability of all the devices by setting $E$ to be 1 (i.e., all the devices run at most one local epoch), and impose systems heterogeneity in a similar way. We show training loss in Figure \ref{fig:loss_full_e1} and testing accuracy in Figure \ref{fig:accuracy_full_e1} in the appendix. Even in these settings, allowing for partial work can  improve convergence compared with \fedavg. Second, we explore a setting without any statistical heterogeneity using an identically distributed synthetic dataset  (Synthetic IID). In this IID setting, as shown in Figure \ref{fig: iid} in Appendix \ref{appen:complete_result}, \fedavg is rather robust under device failure, and tolerating variable amounts of local work may not cause major improvement. This serves as an additional motivation to rigorously study the effect of statistical heterogeneity on new methods designed for federated learning, as simply relying on IID data (a setting unlikely to occur in practice) may not tell a complete story.

\subsection{Statistical Heterogeneity:  Proximal Term}\label{sec: exp_statistical}
\label{section: property_fedprox}

To better understand how the proximal term can be beneficial in heterogeneous settings, we first show convergence can become worse as statistical heterogeneity increases. 

\subsubsection{Effects of Statistical Heterogeneity}\label{sec: exp_synthetic_statistical}

In Figure \ref{fig: synthetic_statistical} (the first row), we study how statistical heterogeneity affects convergence using four synthetic datasets without the presence of systems heterogeneity (fixing $E$ to be 20). From left to right, as data become more heterogeneous, convergence becomes worse for \fedprox with $\mu=0$ (i.e., \fedavg). Though it may slow convergence for IID data, we see that setting $\mu>0$ is particularly useful in heterogeneous settings. This indicates that the modified subproblem introduced in \fedprox can benefit practical federated settings with varying statistical heterogeneity. For perfectly IID data, some heuristics such as decreasing $\mu$ if the loss continues to decrease may help avoid the deceleration of convergence (see Figure \ref{fig: dynamic_mu} in Appendix \ref{app: mu}). In the sections to follow, we see similar results in our non-synthetic experiments.

\subsubsection{Effects of $\mu>0$}
The key parameters of \fedprox that affect performance are the amount of local work (as parameterized by the number of local epochs, $E$), and the proximal term scaled by $\mu$. Intuitively, large $E$ may cause local models to drift too far away from the initial starting point, thus leading to potential divergence~\cite{mcmahan2016FedAvg}. Therefore, to handle the divergence or instability of \fedavg with non-IID data, it is helpful to tune $E$ carefully. However, $E$ is constrained by the underlying system's environments on the devices, and it is difficult to determine an appropriate uniform $E$ for all devices. Alternatively, it is beneficial to allow for device-specific $E$'s (variable $\gamma$'s) and tune a best $\mu$ (a parameter that can be viewed as a re-parameterization of $E$) to prevent divergence and improve the stability of methods. A proper $\mu$ can restrict the trajectory of the iterates by constraining the iterates to be closer to that of the global model, thus incorporating variable amounts of updates and guaranteeing convergence (Theorem \ref{coro: conv_FEDPROX}). 

We show the effects of the proximal term in \fedprox ($\mu>0$) in Figure \ref{fig:loss_all_e20}. For each experiment, we compare the results between \fedprox with $\mu=0$ and \fedprox with a best $\mu$ (see the next paragraph for discussions on how to select $\mu$). For all datasets, we observe that the appropriate $\mu$ can increase the stability for unstable methods and can force divergent methods to converge. This holds both when there is systems heterogeneity (50\% and 90\% stragglers) and there is no systems heterogeneity (0\% stragglers). $\mu>0$ also increases the accuracy in most cases (see Figure \ref{fig: non-iidness_all} and Figure \ref{fig:accuracy_full_e20} in Appendix \ref{appen:complete_result}). In particular, \fedprox improves absolute testing accuracy relative to \fedavg by 22\% on average in highly heterogeneous environments (90\% stragglers) (see Figure \ref{fig:accuracy_full_e20}). 

\begin{figure}
    \centering
    \includegraphics[width=0.42\textwidth,trim={3mm 10mm 3mm 7mm},clip]{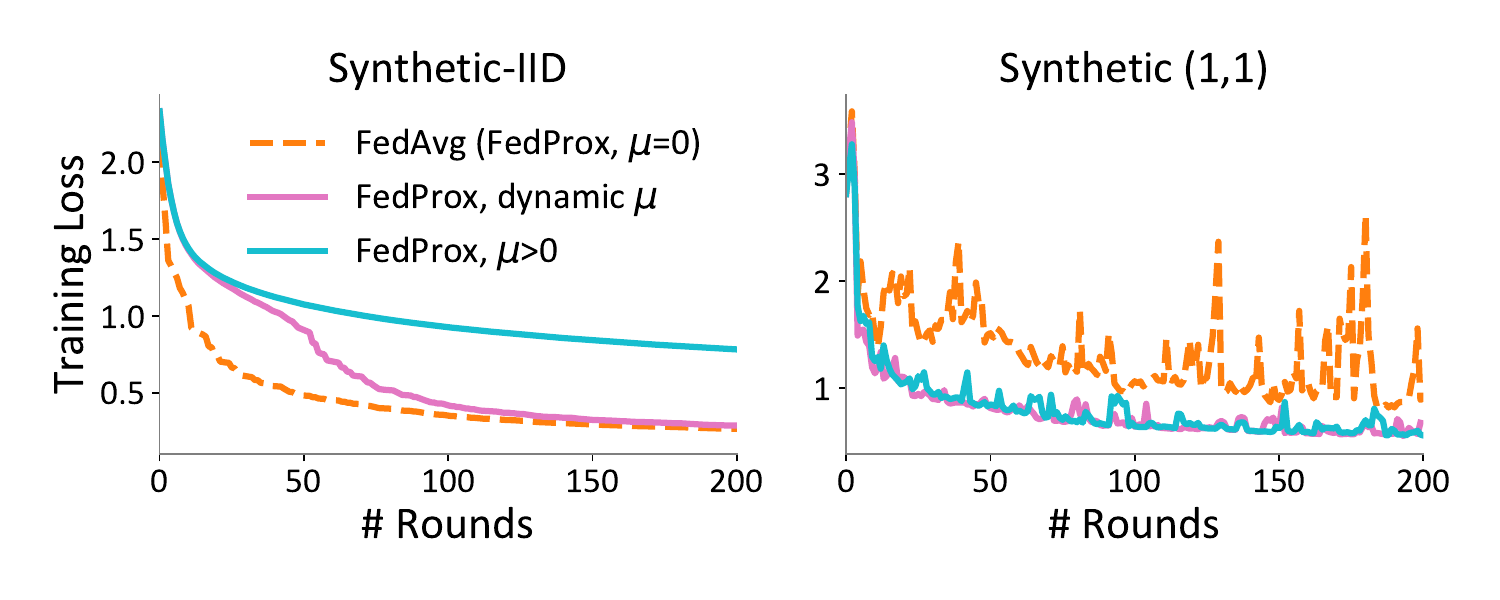}
    \vspace{-2mm}
    \caption{Effectiveness of setting $\mu$ adaptively based on the current model performance. We increase $\mu$ by 0.1 whenever the loss increases and decreases it by 0.1 whenever the loss decreases for 5 consecutive rounds. We initialize $\mu$ to 1 for Synthetic IID (in order to be adversarial to our methods), and initialize $\mu$ to 0 for Synthetic (1,1). This simple heuristic works well empirically.}
    \vspace{1em}
    \label{fig: dynamic_mu_subset}
    
\end{figure}

\vspace{4em}
\textbf{Choosing $\mu$.} One natural question is to determine how to set the penalty constant $\mu$ in the proximal term. A large $\mu$ may potentially slow the convergence by forcing the updates to be close to the starting point, while a small $\mu$ may not make any difference. In all experiments, we tune the best $\mu$ from the limited candidate set $\{0.001, 0.01, 0.1, 1\}$. For the five federated datasets in Figure \ref{fig:loss_all_e20}, the best $\mu$ values are 1, 1, 1, 0.001, and 0.01, respectively. While automatically tuning $\mu$ is difficult to instantiate directly from our theoretical results, in practice, we note that $\mu$ can be adaptively chosen based on the current performance of the model. For example, one simple heuristic is to increase $\mu$ when seeing the loss increasing and decreasing $\mu$ when seeing the loss decreasing. In Figure \ref{fig: dynamic_mu_subset}, we demonstrate the effectiveness of this heuristic using two synthetic datasets. Note that we start from initial $\mu$ values that are adversarial to our methods. We provide full results showing the competitive performance of this approach in Appendix \ref{app: mu}.
Future work includes developing methods to automatically tune this parameter for heterogeneous datasets, based, e.g., on the theoretical groundwork provided here.

\subsubsection{Dissimilarity Measurement and Divergence}\label{sec: exp_dissimilar}

\label{section: sim}
Finally, in Figure \ref{fig: synthetic_statistical} (the bottom row), we demonstrate that our B-local dissimilarity measurement in Definition~\ref{def: similarity} captures the heterogeneity of datasets and is therefore an appropriate proxy of performance.
In particular, we track the variance of gradients on each device, $E_k[\|\nabla F_k(w) - \nabla f(w)\|^2]$, which is lower bounded by $B_\epsilon$ (see Bounded Variance Equivalence Corollary \ref{coro: bounded_var_sim}). 
Empirically, we observe that increasing $\mu$ leads to smaller dissimilarity among local functions $F_k$, and that the dissimilarity metric is consistent with the training loss. Therefore, smaller dissimilarity indicates better convergence, which can be enforced by setting $\mu$ appropriately. We also show the dissimilarity metric on real federated data in Appendix \ref{appen:complete_result}.

\vspace{-2mm}
\section{Conclusion}
\vspace{-2mm}
In this work, we have proposed \fedprox, an optimization framework that tackles the systems and statistical heterogeneity inherent in federated networks. \fedprox allows for variable amounts of work to be performed locally across devices, and relies on a proximal term to help stabilize the method. We provide the convergence guarantees for \fedprox in realistic federated settings under a  device dissimilarity assumption, while also accounting for practical issues such as stragglers. Our empirical evaluation across a suite of federated datasets has validated our theoretical analysis and demonstrated that the \fedprox framework can significantly improve the convergence behavior of federated learning in realistic heterogeneous networks.

\section*{Acknowledgements}

We thank Sebastian Caldas, Jakub Kone\v{c}n\'y, Brendan McMahan, Nathan Srebro, and Jianyu Wang for their helpful discussions. AT and VS are supported in part by DARPA FA875017C0141, the National Science Foundation
grants IIS1705121 and IIS1838017, an Okawa Grant, a Google Faculty Award, an Amazon Web
Services Award,  a JP Morgan A.I. Research Faculty Award, a Carnegie Bosch Institute Research Award, and the CONIX Research Center, one of
six centers in JUMP, a Semiconductor Research Corporation (SRC) program sponsored by DARPA.
Any opinions, findings, and conclusions or recommendations expressed in this material are those of
the author(s) and do not necessarily reflect the views of DARPA, the National Science Foundation, or
any other funding agency.

\bibliography{ref}
\bibliographystyle{mlsys2020}

\appendix
\onecolumn
\section{Complete Proofs}
\subsection{Proof of Theorem \ref{lemma: decrease_0}}
\label{sec:main_res_proof}
\begin{proof}
Using our notion of $\gamma$-inexactness for each local solver (Definition~\ref{def: inexactness}), we can define $e_k^{t+1}$ such that:
	{\small\begin{align}
		&\nabla F_k(w_k^{t+1}) + \mu (w_k^{t+1}-w^{t}) - e_k^{t+1} = 0,\nonumber\\ &\|e_k^{t+1}\|\leq \gamma \|\nabla F_k(w^{t})\| \, .
		\end{align}}
	\noindent Now let us define $\bar{w}^{t+1} = \E{k}{w_k^{t+1}}$. Based on this definition, we know
	{\small\begin{align}
		\bar{w}^{t+1} - w^t = \frac{-1}{\mu} \E{k}{\nabla F_k(w_k^{t+1})} + \frac{1}{\mu}\E{k}{e_k^{t+1}}.
		\end{align}}
	\noindent Let us define $\bar{\mu} = \mu - L_->0$ and $\hat{w}_k^{t+1} = \arg\min_w~h_k(w; w^{t})$. Then, due to the $\bar{\mu}$-strong convexity of $h_k$, we have
	{\small\begin{align}
		\|\hat{w}_k^{t+1}-w_k^{t+1}\|\leq \frac{\gamma}{\bar{\mu}}\|\nabla F_k(w^t)\|.
		\end{align}}
	Note that once again, due to the $\bar{\mu}$-strong convexity of $h_k$, we know that $\|\hat{w}_k^{t+1}- w^t\|\leq \frac{1}{\bar{\mu}}\|\nabla F_k(w^t)\|$. Now we can use the triangle inequality to get 
	{\small\begin{align}
		\|w_k^{t+1}-w^{t}\|\leq \frac{1+\gamma}{\bar{\mu}}\|\nabla F_k(w^{t})\|. \label{eq: bound_0}
		\end{align}}
	Therefore,
	{\small\begin{align}
		&\|\bar{w}^{t+1}-w^{t}\| \leq \E{k}{ \|w_k^{t+1}-w^{t}\|} \leq \frac{1+\gamma}{\bar{\mu}}\E{k}{\|\nabla F_k(w^{t})\|}\nonumber \leq \frac{1+\gamma}{\bar{\mu}}\sqrt{\E{k}{\|\nabla F_k(w^{t})\|^2}}\leq \frac{B(1+\gamma)}{\bar{\mu}}\|\nabla f(w^{t})\|,\label{eq: bound1}
		\end{align}}
	where the last inequality is due to the bounded dissimilarity assumption. 
	
	Now let us define $M_{t+1}$ such that {\small$\bar{w}^{t+1}-w^t = \frac{-1}{\mu} \big( \nabla f(w^t) + M_{t+1}\big)$}, i.e.
$M_{t+1} = \E{k}{\nabla F_k(w_k^{t+1})-\nabla F_k(w^t)-e_k^{t+1}}$.
	We can bound $\|M_{t+1}\|$:
	{\small\begin{align}
		\|M_{t+1}\| & \leq \E{k}{L\|w_k^{t+1} - w_k^t\| + \|e_k^{t+1}\|}\le\Bigg(\frac{L(1+\gamma)}{\bar{\mu}}+ \gamma\Bigg)\times\E{k}{ \|\nabla F_k(w^t)\|}  \leq \Bigg(\frac{L(1+\gamma)}{\bar{\mu}}+ \gamma\Bigg)B\|\nabla f(w^t)\| \, ,
		\end{align}}
	where the last inequality is also due to bounded dissimilarity assumption.
	Based on the L-Lipschitz smoothness of $f$ and Taylor expansion, we have
	{\small\begin{align}
		&f(\bar{w}^{t+1})\leq f(w^t) + \langle \nabla f(w^t), \bar{w}^{t+1}-w^t\rangle + \frac{L}{2}\|\bar{w}^{t+1}-w^t\|^2\nonumber\\
		&\leq f(w^t) -\frac{1}{\mu} \|\nabla f(w^t)\|^2 -\frac{1}{\mu} \langle\nabla f(w^t), M_{t+1}\rangle \nonumber+ \frac{L(1+\gamma)^2B^2}{2\bar{\mu}^2}\|\nabla f(w^t)\|^2
		\nonumber \\
		&\leq f(w^t)-\left(\frac{1-\gamma B}{\mu} - \frac{LB(1+\gamma)}{\bar{\mu}\mu} - \frac{L(1+\gamma)^2B^2}{2\bar{\mu}^2} \right)\times\|\nabla f(w^t)\|^2.\label{eq: bound3}
		\end{align}}
	From the above inequality it follows that if we set the penalty parameter $\mu$ large enough, we can get a decrease in the objective value of $f(\bar{w}^{t+1})-f(w^{t})$ which is proportional to $\|\nabla f(w^t)\|^2$. However, this is not the way that the algorithm works. In the algorithm, we only use $K$ devices that are chosen randomly to approximate $\bar{w}^t$. So, in order to find the $\E{}{ f(w^{t+1})}$, we use local Lipschitz continuity of the function $f$.
	{\small\begin{align}
		f(w^{t+1}) \leq f(\bar{w}^{t+1}) + L_0 \|w^{t+1}-\bar{w}^{t+1}\|, \label{eq: bound5}
		\end{align}}
	where $L_0$ is the local Lipschitz continuity constant of function $f$ and we have
	{\small\begin{align}
		&L_0\leq \|\nabla f(w^{t})\| + L\max(\|\bar{w}^{t+1}-w^{t}\|, \|w^{t+1}-w^{t}\|)\nonumber \leq\|\nabla f(w^{t})\| + L(\|\bar{w}^{t+1}-w^{t}\|+ \|w^{t+1}-w^{t}\|).
		\end{align}}
	Therefore, if we take expectation with respect to the choice of devices in round $t$ we need to bound
	{\small\begin{align}
		\E{S_t}{f(w^{t+1})} \leq f(\bar{w}^{t+1}) + Q_{t},\label{eq: bound2}
		\end{align}}
	where $Q_t = \E{S_t}{L_0 \|w^{t+1}-\bar{w}^{t+1}\|}$. Note that the expectation is taken over the random choice of devices to update.
	{\small\begin{align}
		Q_t &\leq \bE_{S_t}\bigg[\bigg(\|\nabla f(w^{t})\| + L(\|\bar{w}^{t+1}-w^{t}\|+ \|w^{t+1}-w^{t}\|)\bigg)\times\|w^{t+1}-\bar{w}^{t+1}\|\bigg] \nonumber\\
		&\leq \bigg(\|\nabla f(w^{t})\| + L\|\bar{w}^{t+1}-w^{t}\|\bigg)\E{S_t}{ \|w^{t+1}-\bar{w}^{t+1}\|}+ L\E{S_t}{\|w^{t+1}-w^{t}\|\cdot\|w^{t+1}-\bar{w}^{t+1}\|}\nonumber\\
		&\leq \bigg(\|\nabla f(w^{t})\| + 2L\|\bar{w}^{t+1}-w^{t}\|\bigg)\E{S_t}{ \|w^{t+1}-\bar{w}^{t+1}\|}+ L\E{S_t}{\|w^{t+1}-\bar{w}^{t+1}\|^2} \label{eq: S_t}
		\end{align}}
	From \eqref{eq: bound1}, we have that $\|\bar{w}^{t+1}-w^{t}\|\leq \frac{B(1+\gamma)}{\bar{\mu}} \|\nabla f(w^{t})\|$. Moreover,
	{\small\begin{align}
		\E{S_t}{\|w^{t+1}-\bar{w}^{t+1}\|}\leq \sqrt{\E{S_t}{\|w^{t+1}-\bar{w}^{t+1}\|^2}}
		\end{align}}
	and
	{\small\begin{align}
		&\E{S_t}{\|w^{t+1}-\bar{w}^{t+1}\|^2}\leq \frac{1}{K}\E{k}{\|w_k^{t+1}-\bar{w}^{t+1}\|^2}\nonumber\\
		&\leq\frac{2}{K}\E{k}{\|w_k^{t+1}-w^{t}\|^2}, ~~~\text{(as $\bar{w}^{t+1} = \E{k}{w_k^{t+1}}$)}\nonumber\\
		& \leq \frac{2}{K}\frac{(1+\gamma)^2}{\bar{\mu}^2}\E{k}{\|\nabla F_k(w^t)\|^2} ~~~~\text{(from \eqref{eq: bound_0})}\nonumber\\
		& \leq \frac{2B^2}{K}\frac{(1+\gamma)^2}{\bar{\mu}^2}\|\nabla f(w^t)\|^2,
		\end{align}}
	where the first inequality is a result of $K$ devices being chosen randomly to get $w^t$ and the last inequality is due to bounded dissimilarity assumption.
	If we replace these bounds in \eqref{eq: S_t} we get
	{\small\begin{align}
		Q_t\leq \Bigg(\frac{B(1+\gamma)\sqrt{2}}{\bar{\mu}\sqrt{K}}+ \frac{LB^2(1+\gamma)^2}{\bar{\mu}^2 K}\bigg({2\sqrt{2K}+2}\bigg)\Bigg)\|\nabla f(w^{t})\|^2 \label{eq: bound4}
		\end{align}}
	Combining \eqref{eq: bound3}, \eqref{eq: bound2}, \eqref{eq: bound5} and \eqref{eq: bound4} and using the notation $\alpha = \frac{1}{\mu}$ we get
	{\small\begin{align*}
		&\E{S_t}{f(w^{t+1})} \leq f(w^{t}) - \Bigg(\frac{1}{\mu} - \frac{\gamma B}{\mu}-\frac{B(1+\gamma)\sqrt{2}}{\bar{\mu}\sqrt{K}} - \frac{LB(1+\gamma)}{\bar{\mu}\mu}\nonumber\\& - \frac{L(1+\gamma)^2B^2}{2\bar{\mu}^2} - \frac{LB^2(1+\gamma)^2}{\bar{\mu}^2 K}\bigg({2\sqrt{2K}+2}\bigg) \Bigg)\|\nabla f(w^t)\|^2. 
		\end{align*}}
\end{proof}

\subsection{Proof for Bounded Variance}\label{app: boundedvar}

\begin{corollary}[Bounded variance equivalence]\label{coro: bounded_var_sim}
Let Assumption~\ref{assumption: 1} hold.
Then, in the case of bounded variance, i.e., $\E{k}{\|\nabla F_k(w) - \nabla f(w)\|^2}\leq \sigma^2$, for any $\epsilon>0$ it follows that $B_\epsilon \leq \sqrt{1+\frac{\sigma^2}{\epsilon}}$.
\end{corollary}
\paragraph{Proof.} We have,
\begin{align*}
   &E_k[\|\nabla F_k(w) - \nabla f(w)\|^2]=E_k[\|\nabla F_k(w)\|^{2}] - \|\nabla f(w)\|^2\leq \sigma^2\nonumber\\
    &\Rightarrow E_k[\|\nabla F_k(w)\|^{2}] \le \sigma^2+\|\nabla f(w)\|^2\nonumber\\
   &\Rightarrow B_{\epsilon}=\sqrt{\frac{E_k[\|\nabla F_k(w)\|^{2}]}{\|\nabla f(w)\|^2}} \le \sqrt{1+\frac{\sigma^2}{\epsilon}}.
\end{align*}

With Corollary \ref{coro: bounded_var_sim} in place, we can restate the main result in Theorem \ref{lemma: decrease_0} in terms of the bounded variance assumption.

\begin{theorem}[Non-convex \fedprox convergence: Bounded variance]\label{lemma: decrease_1}
Let the assertions of Theorem \ref{lemma: decrease_0} hold. In addition, let the iterate $w^{t}$ be such that $\left\|\nabla f(w^{t})\right\|^{2}\geq\epsilon$, and let $\E{k}{\|\nabla F_k(w) - \nabla f(w)\|^2}\leq \sigma^2$ hold instead of the dissimilarity condition.
If $\mu$, $K$ and $\gamma$ in Algorithm \ref{alg:FEDPROX} are chosen such that
\medmuskip=0mu
\thinmuskip=0mu
\thickmuskip=0mu
\begin{align*}
&\rho = \left(\frac{1}{\mu} - \left(\frac{\gamma }{\mu}+\frac{(1+\gamma)\sqrt{2}}{\bar{\mu}\sqrt{K}} + \frac{L(1+\gamma)}{\bar{\mu}\mu}\right)\sqrt{1+\frac{\sigma^2}{\epsilon}} \right.\nonumber \left.- \left(\frac{L(1+\gamma)^2}{2\bar{\mu}^2} + \frac{L(1+\gamma)^2}{\bar{\mu}^2 K}\bigg({2\sqrt{2K}+2}\bigg)\right)\left(1+\frac{\sigma^2}{\epsilon}\right) \right)>0, \label{eq: rho2}
\end{align*}
then at iteration $t$ of Algorithm \ref{alg:FEDPROX}, we have the following expected decrease in the global objective: 
$$\E{S_t}{ f(w^{t+1})} \leq f(w^{t}) - \rho \|\nabla f(w^t)\|^2 \,,$$ where $S_t$ is the set of $K$ devices chosen at iteration $t$.
\end{theorem}

The proof of Theorem \ref{lemma: decrease_1} follows from the proof of Theorem \ref{lemma: decrease_0} by noting the relationship between the bounded variance assumption and the dissimilarity assumption as portrayed by Corollary \ref{coro: bounded_var_sim}. 

\subsection{Proof of Corollary \ref{coro: convex_exact_fedprox}}
\label{app: convexfedprox}
In the convex case, where $L_- = 0$ and $\bar{\mu} = \mu$, if $\gamma=0$, i.e., all subproblems are solved accurately, we can get a decrease proportional to $\|\nabla f(w^{t})\|^2$ if $B<\sqrt{K}$. 
	In such a case if we assume $1<<B\leq 0.5\sqrt{K}$, then we can write
	{\small\begin{align}
		\E{S_t}{f(w^{t+1})} \lessapprox f(w^{t}) -\frac{1}{2\mu} \|\nabla f(w^{t})\|^2 + \frac{3LB^2}{2\mu^2}\|\nabla f(w^{t})\|^2 \, .
		\end{align}}
	In this case, if we choose ${\mu}\approx {6LB^2}$ we get
	{\small\begin{align}
	\label{eq:final_step}
		\E{S_t}{f(w^{t+1})} \lessapprox f(w^{t}) - \frac{1}{24LB^2}\|\nabla f(w^{t})\|^2 \, .
		\end{align}}
Note that the expectation in \eqref{eq:final_step} is a conditional expectation conditioned on the previous iterate. Taking expectation of both sides, and telescoping, we have that the number of iterations to at least generate one solution with squared norm of gradient less than $\epsilon$ is $O(\frac{LB^2\Delta}{\epsilon})$.

\newpage
\section{Connections to other single-machine and distributed methods}
\label{app: relwork}

Two aspects of the proposed work---the proximal term in \fedprox, and the bounded dissimilarity assumption used in our analysis---have been previously studied in the optimization literature, but with very different motivations. For completeness, we provide a discussion below on our relation to these prior works.

\paragraph{Proximal term.}

The proposed modified objective in \fedprox shares a connection with elastic averaging SGD (EASGD)~\cite{elastic_SGD_zhang_LeCun_2015}, which was proposed as a way to train deep networks in the data center setting, and uses a similar proximal term in its objective. While the intuition is similar to EASGD (this term helps to prevent large deviations on each device/machine), EASGD employs a more complex moving average to update parameters, is limited to using SGD as a local solver, and has only been analyzed for simple quadratic problems. The proximal term we introduce has also been explored in previous optimization literature with  different purposes, such as~\citet{allen2018make}, to speed up (mini-batch) SGD training on a single machine, and in \citet{li2014efficient} for efficient SGD training both in a single machine and distributed settings. However, the analysis in~\citet{li2014efficient} is limited to a single machine setting with different assumptions (e.g., IID data and solving the subproblem exactly at each round). 

\begin{figure*}[h]
    \centering
    \begin{subfigure}{1\textwidth}
    \includegraphics[width=\textwidth]{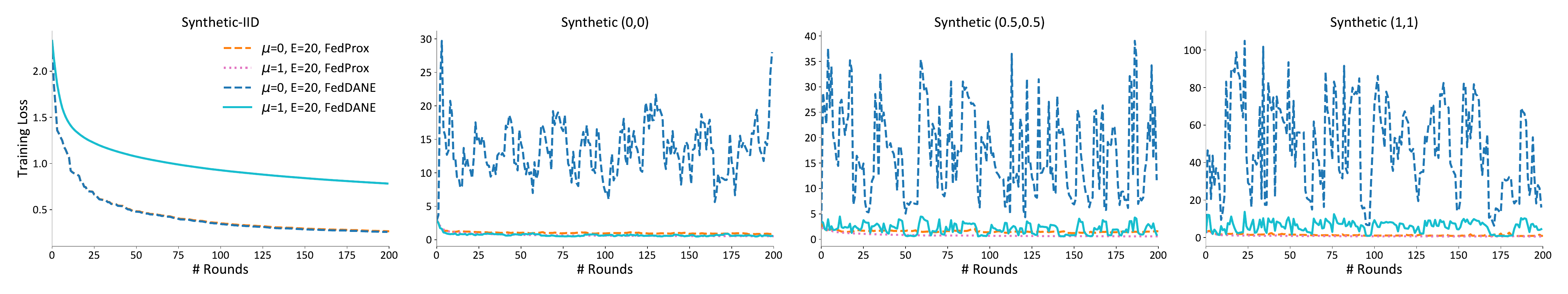}
    \end{subfigure}
    
    \begin{subfigure}{1\textwidth}
    \includegraphics[width=\textwidth]{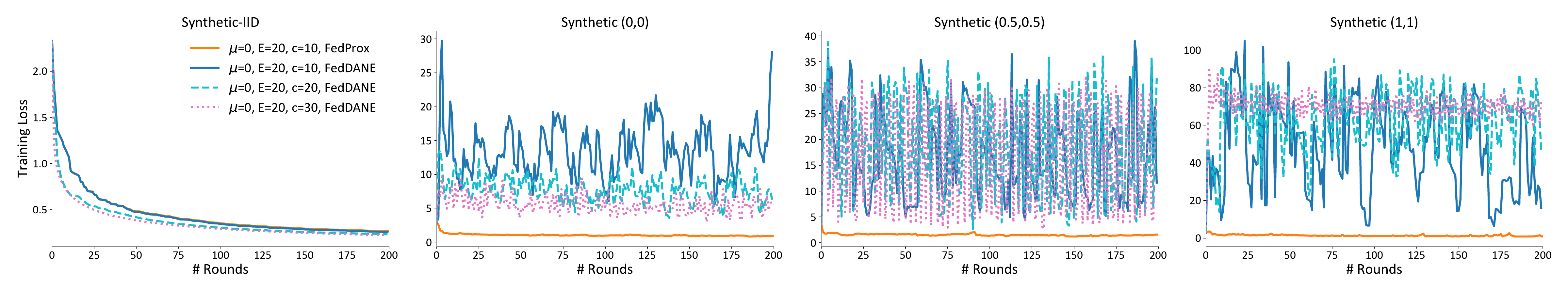}
    \end{subfigure}
    
    \caption{DANE and AIDE~\cite{shamir2014communication, AIDE_reddi_16} are methods proposed in the data center setting that use a similar proximal term as \fedprox as well as an additional gradient correction term. We modify DANE to apply to federated settings by allowing for local updating and low participation of devices. We show the convergence of this modified method, which we call \feddane, on synthetic datasets. In the top figures, we subsample 10 devices out of 30 on all datasets for both \fedprox and \feddane. While \feddane performs similarly as \fedprox on the IID data, it suffers from poor convergence on the non-IID datasets. In the bottom figures, we show the results of \feddane when we increase the number of selected devices in order to narrow the gap between our estimated full gradient and the real full gradient (in the gradient correction term). Note that communicating with all (or most of the) devices is already unrealistic in practical settings. We observe that although sampling more devices per round might help to some extent, \feddane is still unstable and tends to diverge. This serves as additional motivation for the specific subproblem we propose in \fedprox.}
    \label{fig:feddane}
\end{figure*}

In addition, DANE~\cite{shamir2014communication} and AIDE~\cite{AIDE_reddi_16}, distributed methods designed for the data center setting, propose a similar proximal term in the local objective function, but also augment this with an additional gradient correction term. Both methods assume that all devices participate at each communication round, which is impractical in federated settings. Indeed, due to the inexact estimation of full gradients (i.e., $\nabla \phi(w^{(t-1)})$ in \citet[Eq (13)]{shamir2014communication}) with device subsampling schemes and the staleness of the gradient correction term~\citep[Eq (13)]{shamir2014communication}, these methods are not directly applicable to our setting. Regardless of this, we explore a variant of such an approach in federated settings and see that the gradient direction term does not help in this scenario---performing uniformly worse than the proposed \fedprox framework for heterogeneous datasets, despite the extra computation required (see Figure \ref{fig:feddane}). {We refer interested readers to~\citet{li2020feddane} for more detailed discussions.} 

{Finally, we note that there is an interesting connection between meta-learning methods and federated optimization methods~\cite{khodak2019adaptive}, and similar proximal terms have recently been investigated in the context of meta-learning for improved performance on few-shot learning tasks~\cite{goldblum2020unraveling,zhou2019efficient}.}

\paragraph{Bounded dissimilarity assumption.}

The bounded dissimilarity assumption we discuss in Assumption~\ref{assumption: 1} has appeared in different forms, for example in \citet{SGD_under_growth_schmidt_2013, pmlr-v84-yin18a,vaswani2018fast}. In \citet{pmlr-v84-yin18a}, the bounded similarity assumption is used in the context of asserting gradient diversity and quantifying the benefit in terms of scaling of the mean square error for mini-batch SGD for IID data. In \citet{SGD_under_growth_schmidt_2013, vaswani2018fast}, the authors use a similar assumption, called \emph{strong growth condition}, which is a stronger version of Assumption \ref{assumption: 1} with $\epsilon=0$. They prove that some interesting practical problems satisfy such a condition. They also use this assumption to prove optimal and better convergence rates for SGD with constant step-sizes. Note that this is different from our approach as the algorithm that we are analyzing is not SGD, and our analysis is different in spite of the similarity in the assumptions.

\newpage
\section{Simulation Details and Additional Experiments}
\label{app:exps}

\subsection{Datasets and Models}
\label{app:data}
\begin{figure*}[h]
    \centering
    \begin{subfigure}{1\textwidth}
        \includegraphics[width=\textwidth]{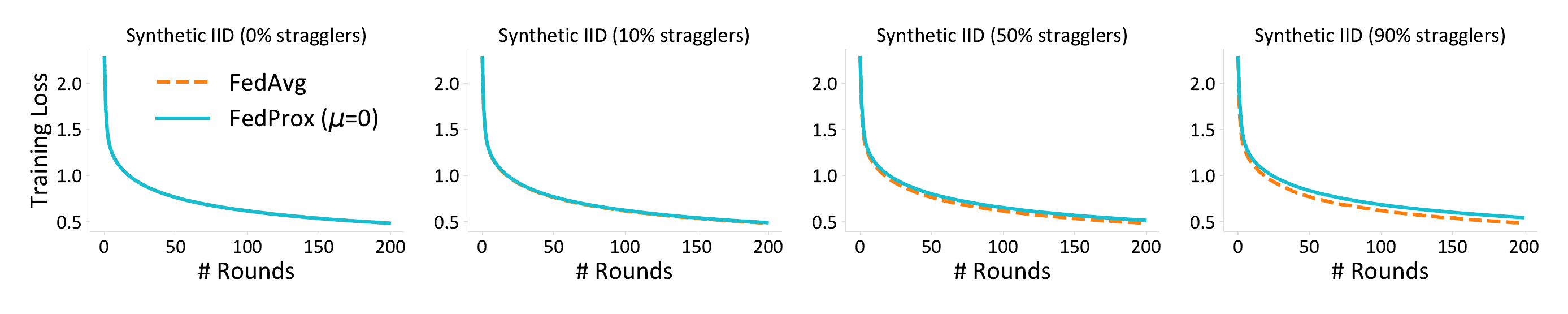}
    \end{subfigure}
    
    \begin{subfigure}{1\textwidth}
        \includegraphics[width=\textwidth]{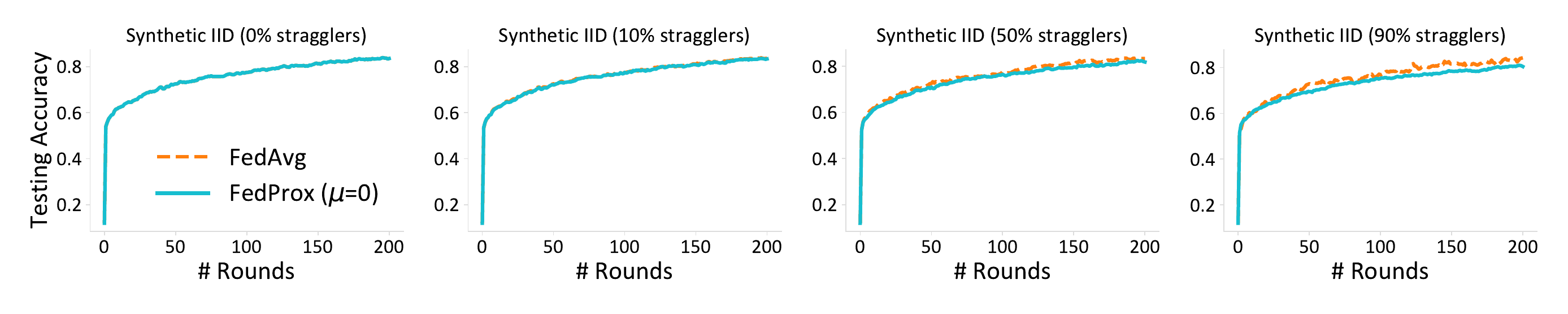}
    \end{subfigure}
    \caption{\fedavg is robust to device failure with IID data. In this case, whether incorporating partial solutions from the stragglers would not have much effect on convergence.}
    \label{fig: iid}

\end{figure*}

Here we provide full details on the datasets and models used in our experiments. We curate a diverse set of non-synthetic datasets, including those used in prior work on federated learning~\cite{mcmahan2016FedAvg}, and some proposed in LEAF, a benchmark for federated settings~\cite{caldas2018leaf}. We also create synthetic data to directly test the effect of heterogeneity on convergence, as in Section~\ref{section: set up}.
\begin{itemize}[leftmargin=*]
\item\textbf{Synthetic:} We set $(\alpha, \beta)$=(0,0), (0.5,0.5) and (1,1) respectively to generate three non-identical distributed datasets (Figure \ref{fig: synthetic_statistical}). In the IID data (Figure \ref{fig: iid}), we set the same $W, b \sim \mathcal{N}(0,1)$ on all devices and $X_k$ to follow the same distribution $\mathcal{N}(v, \Sigma)$ where each element in the mean vector $v$ is zero and $\Sigma$ is diagonal with $\Sigma_{j,j}=j^{-1.2}$. For all synthetic datasets, there are $30$ devices in total and the number of samples on each device follows a power law.

\item\textbf{MNIST:}    We study image classification of handwritten digits 0-9 in MNIST~\cite{lecun1998gradient} using multinomial logistic regression. To simulate a heterogeneous setting, we distribute the data among $1000$ devices such that each device has samples of only $2$ digits and the number of samples per device follows a power law. The input of the model is a flattened 784-dimensional (28 $\times$ 28) image, and the output is a class label between 0 and 9.

\item\textbf{FEMNIST:} We study an image classification problem on the $62$-class EMNIST dataset~\cite{cohen2017emnist} using multinomial logistic regression. To generate heterogeneous data partitions, we subsample $10$ lower case characters (`a'-`j') from EMNIST and distribute only 5 classes to each device. We call this \emph{federated} version of EMNIST \emph{FEMNIST}. There are 200 devices in total. The input of the model is a flattened 784-dimensional (28 $\times$ 28) image, and the output is a class label between 0 and 9.

\item\textbf{Shakespeare:} This is a dataset built from \emph{The Complete Works of William Shakespeare}~\cite{mcmahan2016FedAvg}. Each speaking role in a play represents a different device. We use a two-layer LSTM classifier containing 100 hidden units with an 8D embedding layer.
The task is next-character prediction, and there are 80 classes of characters in total. The model takes as input a sequence of 80 characters, embeds each of the characters into a learned 8-dimensional space and outputs one character per training sample after 2 LSTM layers and a densely-connected layer. 

\item\textbf{Sent140:} In non-convex settings, we consider a text sentiment analysis task on tweets from Sentiment140~\cite{go2009twitter} (Sent140) with a two layer LSTM binary classifier containing 256 hidden units with pretrained 300D GloVe embedding~\citep{pennington2014glove}. 
Each twitter account corresponds to a device. The model takes as input a sequence of 25 characters, embeds each of the characters into a 300-dimensional space by looking up Glove and outputs one character per training sample after 2 LSTM layers and a densely-connected layer. 
\end{itemize}

\subsection{Implementation Details}
\label{app:implementation}
(\textbf{Implementation}) In order to draw a fair comparison with \fedavg, we use SGD as a local solver for \fedprox, and adopt a slightly different device sampling scheme than that in Algorithms~\ref{alg:FEDAVG} and~\ref{alg:FEDPROX}: sampling devices uniformly and averaging updates with weights proportional to the number of local data points (as originally proposed in~\citet{mcmahan2016FedAvg}). While this sampling scheme is not supported by our analysis, we observe similar relative behavior of \fedprox vs. \fedavg whether or not it is employed (Figure \ref{fig: compare_sampling}). Interestingly, we also observe that the sampling scheme proposed herein results in more stable performance for both methods. This suggests an added benefit of the proposed framework.

(\textbf{Machines}) We simulate the federated learning setup (1 server and $N$ devices) on a commodity machine with 2 Intel$^\text{\textregistered}$ Xeon$^\text{\textregistered}$ E5-2650 v4 CPUs and 8 NVidia$^\text{\textregistered}$ 1080Ti GPUs.

(\textbf{Hyperparameters}) We randomly split the data on each local device into an 80\% training set and a 20\% testing set. We fix the number of selected devices per round to be 10 for all experiments on all datasets. 
We also do a grid search on the learning rate based on \fedavg. We do not decay the learning rate through all rounds. For all synthetic data experiments, the learning rate is 0.01. For MNIST, FEMNIST, Shakespeare, and Sent140, we use the learning rates of 0.03, 0.003, 0.8, and 0.3. We use a batch size of 10 for all experiments.

(\textbf{Libraries}) All code is implemented in Tensorflow Version 1.10.1~\cite{tensorflow2015-whitepaper}. Please see \href{https://github.com/litian96/FedProx}{github.com/litian96/FedProx}  for full details.

\subsection{Additional Experiments and Full Results}
\label{app:full experiments}

\subsubsection{Effects of Systems Heterogeneity on IID Data}
We show the effects of allowing for partial work on a perfect IID synthetic data (Synthetic IID). 

\subsubsection{Complete Results} \label{appen:complete_result}

\begin{figure*}[h]
    \centering
    \begin{subfigure}{0.9\textwidth}
        \includegraphics[width=\textwidth]{Figs/non-iidness_loss.pdf}
    \end{subfigure}
    
    \begin{subfigure}{0.9\textwidth}
        \includegraphics[width=\textwidth]{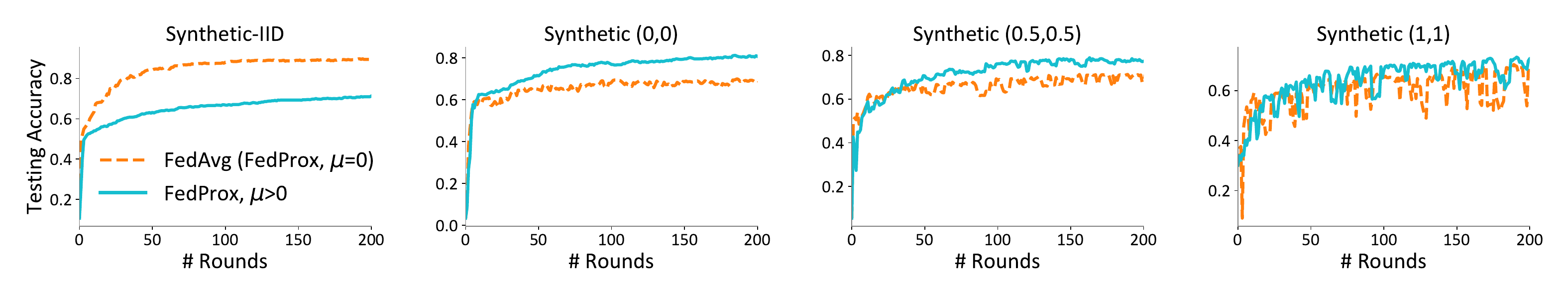}
    \end{subfigure}
    
    \begin{subfigure}{0.9\textwidth}
        \includegraphics[width=\textwidth]{Figs/non-iidness_sim.pdf}
    \end{subfigure}
    \vspace{-3mm}
    \caption{Training loss, test accuracy, and dissimilarity measurement for experiments described in Fig. \ref{fig: synthetic_statistical}.}
    \label{fig: non-iidness_all}
    \vspace{-2mm}
\end{figure*}

In Figure \ref{fig: non-iidness_all}, we present testing accuracy on four synthetic datasets associated with the experiments shown in Figure \ref{fig: synthetic_statistical}.

\begin{figure*}[h]
    \centering
    \includegraphics[width=1\textwidth]{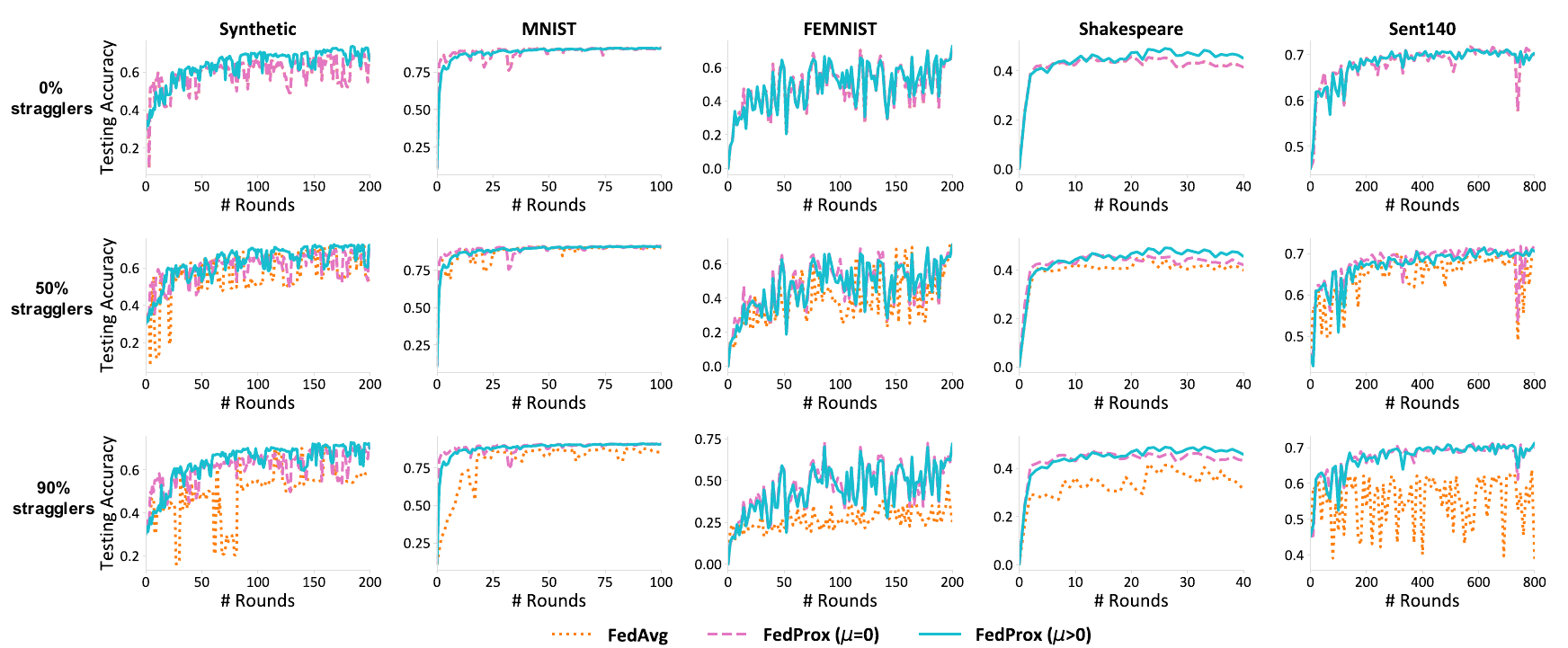}
    \caption{The testing accuracy of the experiments in Figure \ref{fig:loss_all_e20}. \fedprox achieves on average 22\% improvement in terms of testing accuracy in highly heterogeneous settings (90\% stragglers).}
    \label{fig:accuracy_full_e20}
\end{figure*}

\newpage
In Figure \ref{fig:accuracy_full_e20}, we show the testing accuracy associated with the experiments described in Figure \ref{fig:loss_all_e20}. We calculate the accuracy improvement numbers by identifying the accuracies of \fedprox and \fedavg when they have either converged, started to diverge, or run sufficient number of rounds (e.g., 1000 rounds), whichever comes earlier. We consider the methods to converge when the loss difference in two consecutive rounds $|f_t-f_{t-1}|$ is smaller than 0.0001, and consider the methods to diverge when we see $f_{t}-f_{t-10}$ greater than 1. 

\begin{figure}[h]
    \centering
    \includegraphics[width=\textwidth]{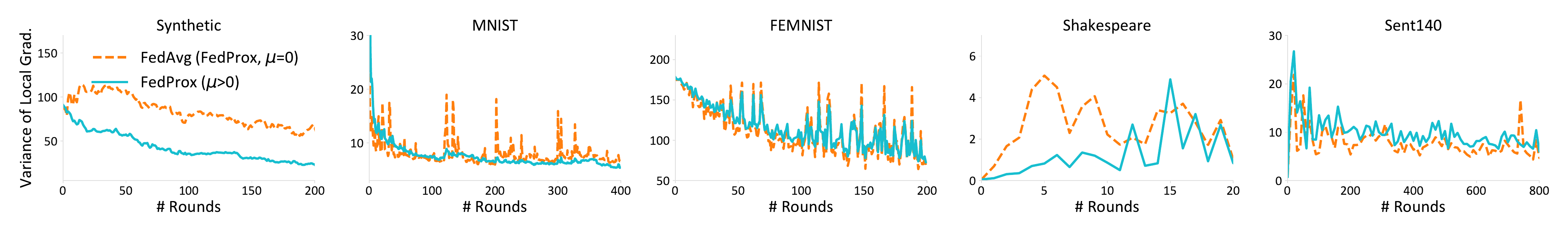}
    \caption{The dissimilarity metric on five datasets in Figure \ref{fig:loss_all_e20}. We remove systems heterogeneity by only considering the case when no participating devices drop out of the network. Our dissimilarity assumption captures the data heterogeneity and is consistent with practical performance (see training loss in Figure \ref{fig:loss_all_e20}).}
    \label{fig:variance_all}
\end{figure}

In Figure \ref{fig:variance_all}, we report the dissimilarity measurement on five datasets (including four real datasets) described in Figure \ref{fig:loss_all_e20}. Again, the dissimilarity characterization is consistent with the real performance (the loss).

\begin{figure*}[h]
    \centering
    \includegraphics[width=1\textwidth]{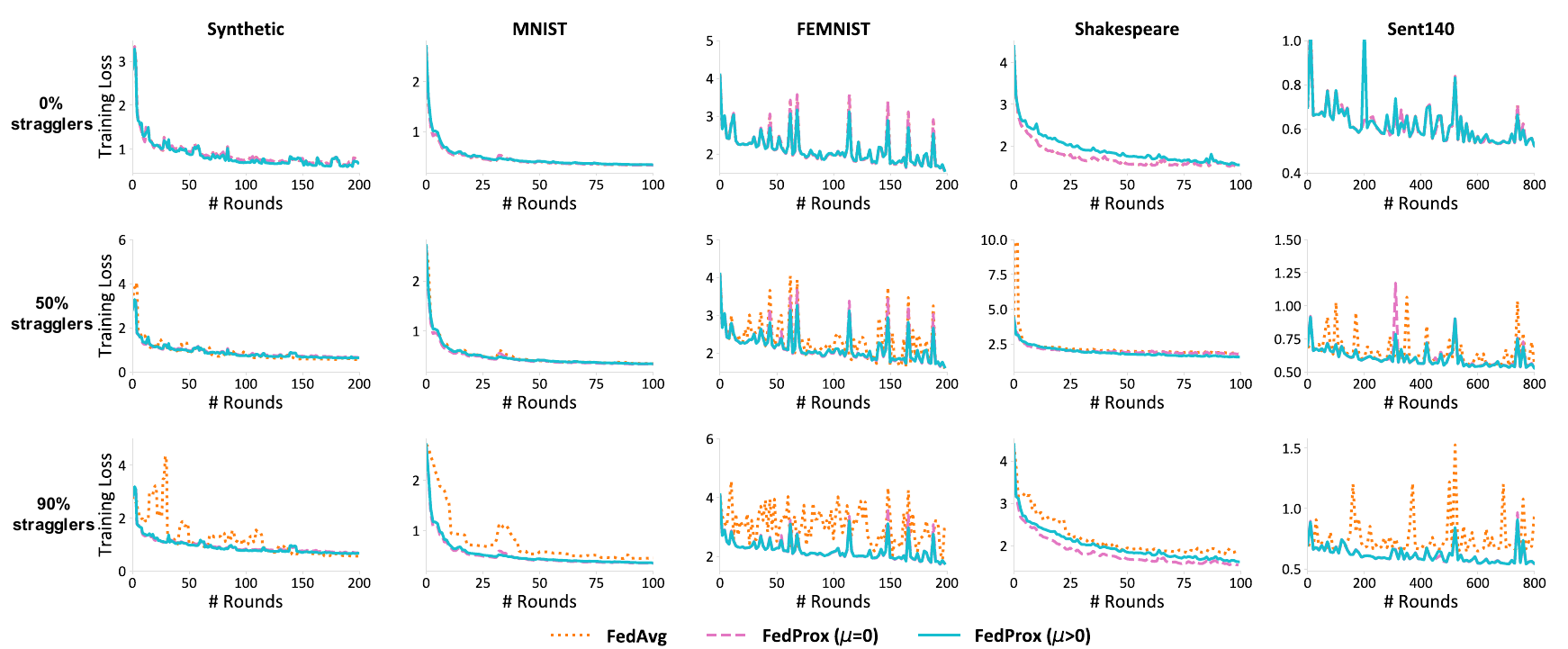}
    \caption{The loss of \fedavg and \fedprox under various systems heterogeneity settings when each device can run at most 1 epoch at each iteration ($E=1$). Since local updates will not deviate too much from the global model compared with the deviation under large $E$'s, it is less likely that the statistical heterogeneity will affect convergence negatively. Tolerating for partial solutions to be sent to the central server (\fedprox, $\mu=0$) still performs better than dropping the stragglers ($\fedavg$).}
    \label{fig:loss_full_e1}
\end{figure*}

\begin{figure*}[h]
    \centering
    \includegraphics[width=1\textwidth]{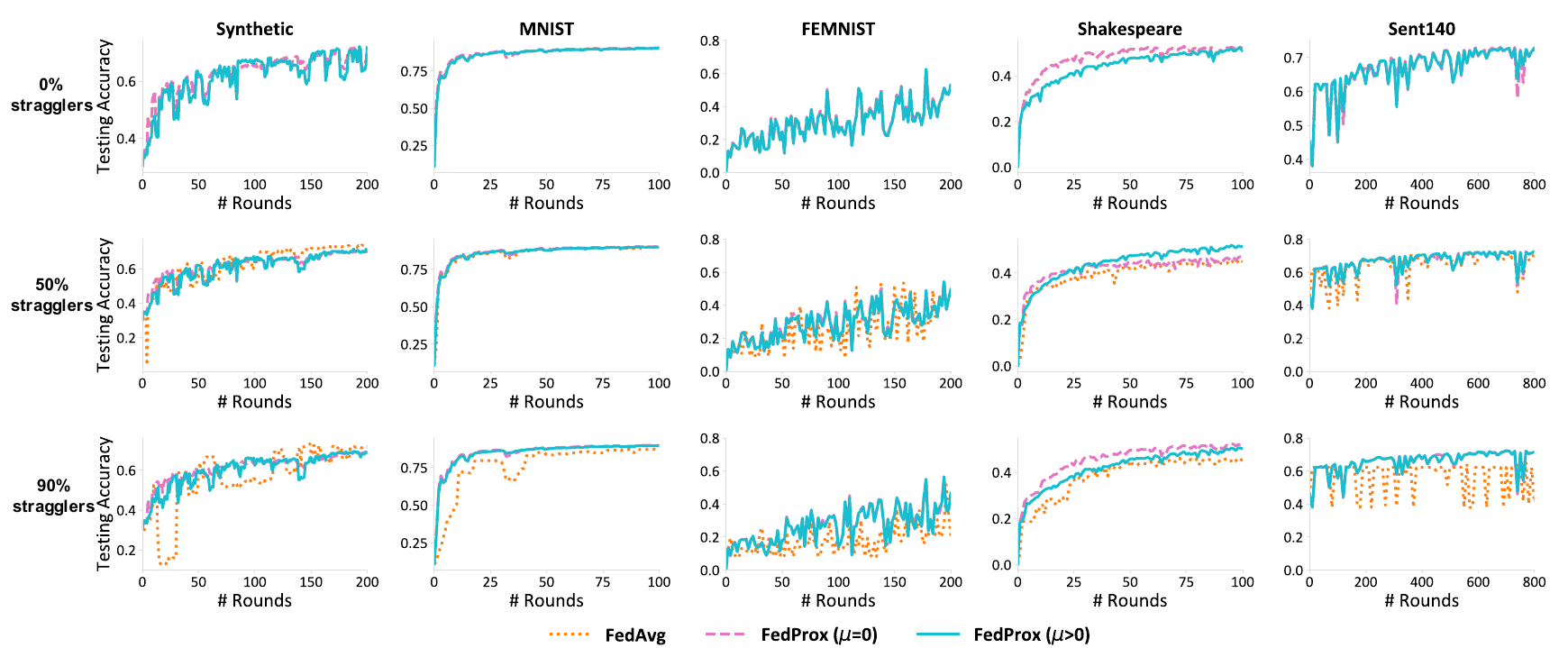}
    \caption{The testing accuracy of the experiments shown in Figure \ref{fig:loss_full_e1}.}
    \label{fig:accuracy_full_e1}
\end{figure*}

\newpage
In Figure \ref{fig:loss_full_e1} and Figure \ref{fig:accuracy_full_e1}, we show the effects  (both loss and testing accuracy) of allowing for partial solutions under systems heterogeneity when $E=1$ (i.e., the statistical heterogeneity is less likely to affect convergence negatively).

\subsubsection{Adaptively setting $\mu$}
\label{app: mu}

\begin{figure}[h]
    \centering
    \includegraphics[width=\textwidth]{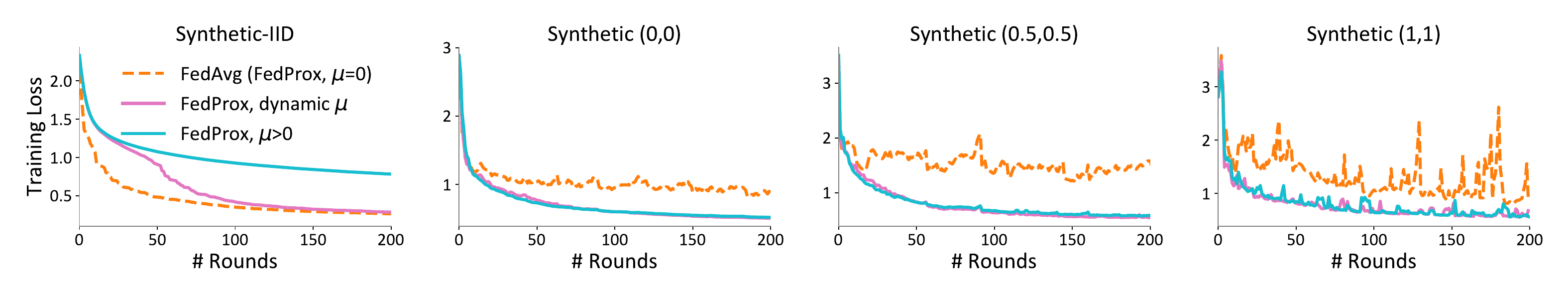}
    \caption{Full results of choosing $\mu$ adaptively on all the synthetic datasets. We increase $\mu$ by 0.1 whenever the loss increases and decreases it by 0.1 whenever the loss decreases for 5 consecutive rounds. We initialize $\mu$ to 1 for the IID data (Synthetic-IID) (in order to be adversarial to our methods), and initialize it to 0 for the other three non-IID datasets. We observe that this simple heuristic works well in practice.}
    \label{fig: dynamic_mu}
\end{figure}

One of the key parameters of \fedprox is $\mu$.
We provide the complete results of a simple heuristic of adaptively setting $\mu$ on four synthetic datasets in Figure \ref{fig: dynamic_mu}. For the IID dataset (Synthetic-IID), $\mu$ starts from 1, and for the other non-IID datasets, $\mu$ starts from 0. Such initialization is adversarial to our methods. We decrease $\mu$ by 0.1 when the loss continues to decrease for 5 rounds and increase $\mu$ by 0.1 when we see the loss increase. This heuristic allows for competitive performance. It could also alleviate the potential issue that $\mu>0$ might slow down convergence on IID data, which rarely occurs in real federated settings.

\subsubsection{Comparing Two Device Sampling Schemes}
\label{app:compare_scheme}

\begin{figure*}[h]
    \centering
    \begin{subfigure}{0.95\textwidth}
    \includegraphics[width=\textwidth]{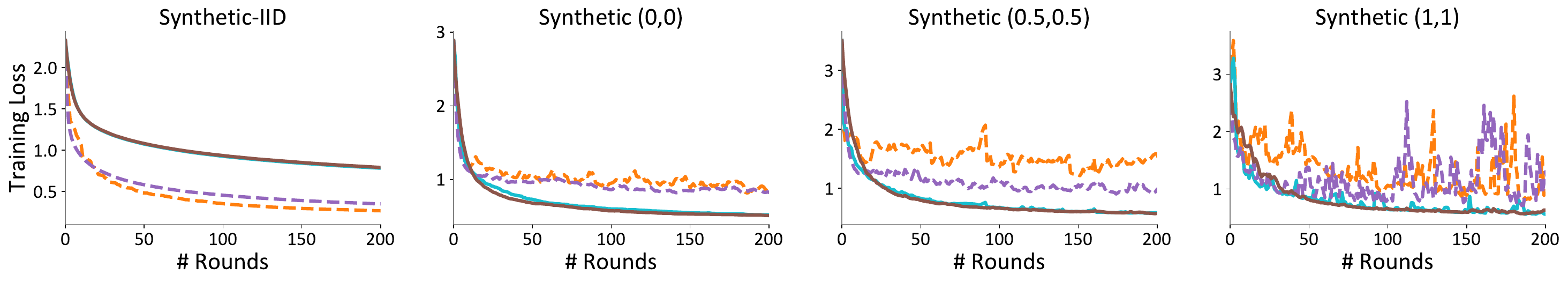}
    \end{subfigure}
    
    \begin{subfigure}{0.95\textwidth}
    \includegraphics[width=\textwidth]{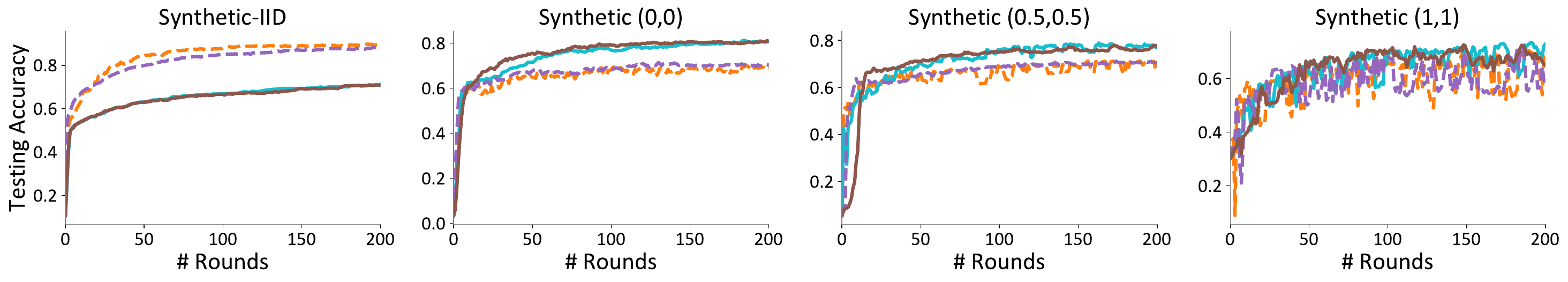}
    \end{subfigure}
    
    \begin{subfigure}{0.95\textwidth}
    \includegraphics[width=\textwidth]{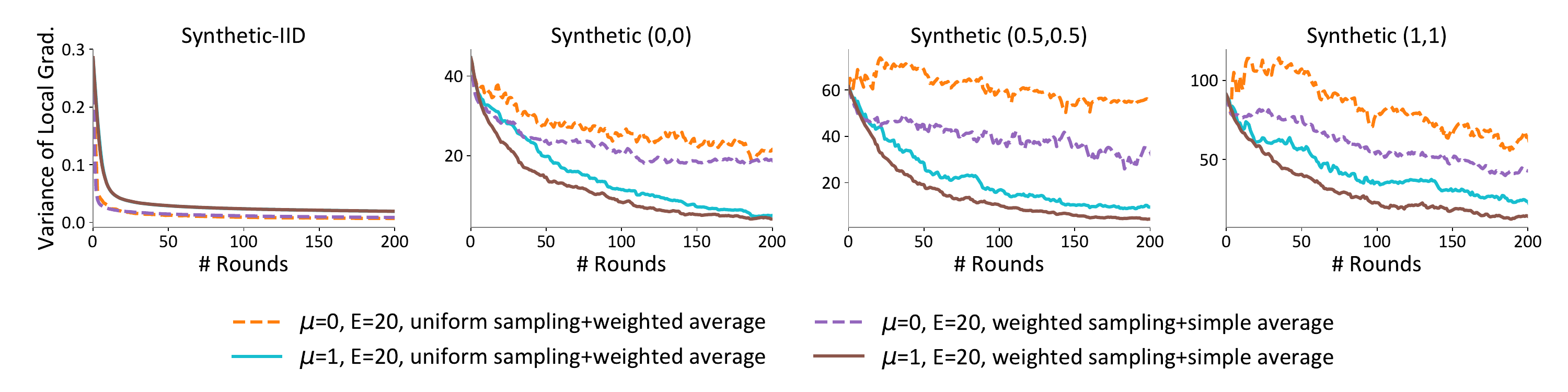}
    \end{subfigure}
    
    \caption{Differences between two sampling schemes in terms of training loss, testing accuracy, and dissimilarity measurement. Sampling devices with a probability proportional to the number of local data points and then simply averaging local models performs slightly better than uniformly sampling devices and averaging the local models with weights proportional to the number of local data points. Under either sampling scheme, the settings with $\mu=1$ demonstrate more stable performance than settings with $\mu=0$.}
    \label{fig: compare_sampling}
\end{figure*}

We show the training loss, testing accuracy, and dissimilarity measurement of \fedprox on a set of synthetic data using two different device sampling schemes in Figure \ref{fig: compare_sampling}. Since our goal is to compare these two sampling schemes, we let each device perform the uniform amount of work ($E=20$) for both methods.

\end{document}